\documentclass[lettersize,journal]{IEEEtran}
\usepackage{amsmath,amsfonts}
\usepackage{algorithmic}
\usepackage{algorithm}
\usepackage{array}
\usepackage{multirow}
\usepackage[caption=false,font=normalsize,labelfont=sf,textfont=sf]{subfig}
\usepackage{textcomp}
\usepackage{stfloats}
\usepackage{url}
\usepackage{xcolor}
\usepackage{verbatim}
\usepackage{graphicx}
\usepackage{cite}
\usepackage{fancyhdr} 

\begin{document}

\title{AVTENet: A Human-Cognition-Inspired Audio-Visual Transformer-Based Ensemble Network for Video Deepfake Detection}

\author{{Ammarah Hashmi, Sahibzada Adil Shahzad, Chia-Wen Lin,~\IEEEmembership{Fellow,~IEEE}, Yu Tsao,~\IEEEmembership{Senior Member,~IEEE}, and Hsin-Min Wang,~\IEEEmembership{Senior Member,~IEEE}
        % <-this % stops a space
\thanks{Ammarah Hashmi is with the Social Networks and Human-Centered Computing Program, Taiwan International Graduate Program, Academia Sinica, Taipei 11529, Taiwan, and also with the Institute of Information Systems and Applications, National Tsing Hua University, Hsinchu 30013, Taiwan. (e-mail: hashmiammarah0@gmail.com).

Sahibzada Adil Shahzad is with the Social Networks and Human-Centered Computing Program, Taiwan International Graduate Program, Academia Sinica, Taipei 11529, Taiwan, and also with the Department of Computer Science, National
Chengchi University, Taipei 11605, Taiwan. (e-mail: adilshah275@iis.sinica.edu.tw).

Chia-Wen Lin is with the Department of Electrical Engineering and the Institute of Communications Engineering, National Tsing Hua University, Hsinchu 300044, Taiwan. (e-mail: cwlin@ee.nthu.edu.tw)

Yu Tsao is with the Research Center for Information Technology Innovation, Academia Sinica, Taipei 11529, Taiwan. (e-mail:
yu.tsao@citi.sinica.edu.tw).

Hsin-Min Wang is with the Institute of Information Science, Academia Sinica, Taipei 11529, Taiwan. (Corresponding Author, e-mail: whm@iis.sinica.edu.tw)}
}}% <-this % stops a space
% \thanks{Manuscript received April 19, 2021; revised August 16, 2021.}

\maketitle
\thispagestyle{fancy} 
\begin{abstract}
The recent proliferation of hyper-realistic deepfake videos has drawn attention to the threat of audio and visual forgeries. Most previous studies on detecting artificial intelligence-generated fake videos only utilize visual modality or audio modality. While some methods exploit audio and visual modalities to detect forged videos, they have not been comprehensively evaluated on multimodal datasets of deepfake videos involving acoustic and visual manipulations, and are mostly based on convolutional neural networks with low detection accuracy. Considering that human cognition instinctively integrates multisensory information including audio and visual cues to perceive and interpret content and the success of transformer in various fields, this study introduces the audio-visual transformer-based ensemble network (AVTENet). This innovative framework tackles the complexities of deepfake technology by integrating both acoustic and visual manipulations to enhance the accuracy of video forgery detection. Specifically, the proposed model integrates several purely transformer-based variants that capture video, audio, and audio-visual salient cues to reach a consensus in prediction. For evaluation, we use the recently released benchmark multimodal audio-video FakeAVCeleb dataset. For a detailed analysis, we evaluate AVTENet, its variants, and several existing methods on multiple test sets of the FakeAVCeleb dataset. Experimental results show that the proposed model outperforms all existing methods and achieves state-of-the-art performance on Testset-I and Testset-II of the FakeAVCeleb dataset. We also compare AVTENet against humans in detecting video forgery. The results show that AVTENet significantly outperforms humans.
\end{abstract}


\section{Introduction}
\IEEEPARstart{W}{idespread} misinformation or disinformation~\cite{R15} on online social media has fueled the spread of unverified rumors, often escalating into social disputes. Deepfake technology further exacerbates this issue. Deepfake refers to forged media content in which the face and/or voice of the source person are manipulated with the face and/or voice of a target person using deep learning algorithms. Deepfake detection is essential for applications such as digital forensic analysis, content moderation, and protecting public trust in media integrity.

\begin{figure}[!t]
\centering
\includegraphics[width=3.28in]{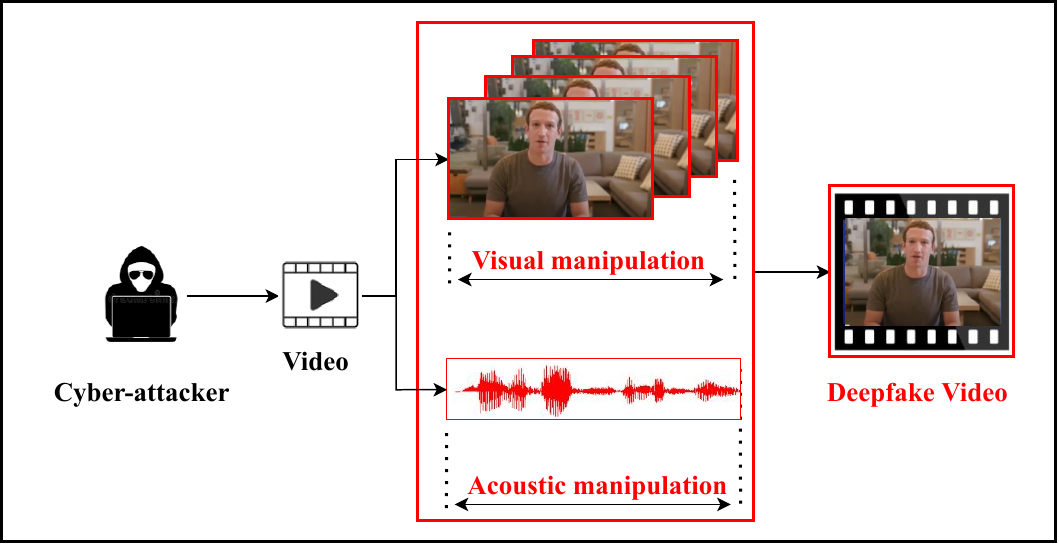}
\vspace{-2mm}
\caption{Example of audio-visual deepfakes. Cyber-attackers aim to create convincing counterfeit videos by manipulating audio and visual streams. Sharing such malicious content can have harmful consequences. Therefore, our proposed method exploits manipulations in both audio and visual streams to effectively detect forged videos.}
\label{Fig1}
\vspace{-6mm}
\end{figure}

Modern deep learning technology has significantly enhanced the generation or manipulation of media content through various advanced tools and techniques. Generative adversarial networks (GANs) \cite{R2} and variational autoencoders (VAEs) \cite{R3} can create highly realistic synthetic media content that is indistinguishable from real content. Face swapping, face reenactment, face editing, and face synthesis are various categories of deepfake techniques that help produce convincing and well-crafted facial deepfake media \cite{R1}. Initially, deepfake content involved unimodal manipulation of audio or
video. But recently, multimodal manipulation has been used to produce more realistic deepfake videos. The viral Instagram video of Mark Zuckerberg (Meta’s CEO), as shown in Fig. 1, is the ultimate example of audio-visual deepfakes that combine acoustic and visual forgeries.

Detecting such convincing deepfake videos is crucial and challenging. To combat the trend of deepfakes, several deep learning-based solutions have been proposed to detect forgeries in videos. Existing deepfake detection methods only focus on unimodal detection of video deepfakes, lacking evaluation on multimodal deepfake datasets. Unimodal methods limit the performance of detectors to certain scenarios; for instance, an audio-only detector (Fig. 2(a)) cannot detect visual manipulation, while a detector that relies solely on the visual modality (Fig. 2(b)) can easily be fooled by acoustic manipulation.

The human brain inherently combines auditory and visual cues to comprehend the world. This multisensory cognitive ability \cite{R72,R73} or multimodal data \cite{R77} brings greater richness and representation and helps identify inconsistencies or anomalies in deepfake media, as shown in Fig. 2(d). Therefore, by drawing inspiration from human cognitive skills and attention mechanisms, our approach bridges the gap between natural and artificial systems, leveraging auditory and visual sensory information \cite{R77} and transformer networks \cite{R39} to detect deepfake videos (Fig. 2(c)), and is evaluated on a multimodal dataset to take a step toward applications in real-world scenarios. 

\begin{figure}[!t]
\centering
\includegraphics[width=2.6in, height=3.9in]{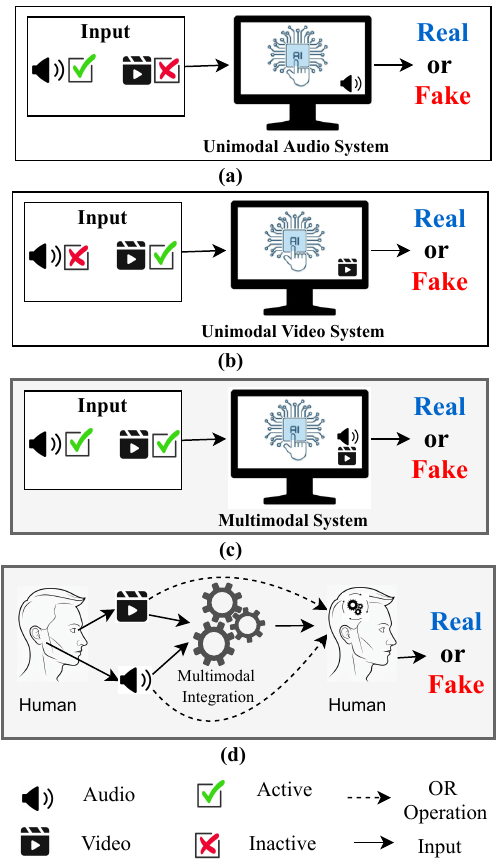}
\vspace{-2mm}
\caption{Comparison of artificial systems and humans in detecting video forgery.
}
\label{newFigure}
\vspace{-6mm}
\end{figure}

Recently, transformers \cite{R39} have emerged as a promising alternative and marked a dominant shift from convolutional neural networks (CNNs) in computer vision tasks due to their ability to model long-range dependencies and capture global context. The sequential processing nature of CNN limits the %network's 
ability of the network to capture long-range dependencies; however, the attention mechanism in the transformer allows it to model long-range global context and relationships. Existing methods mainly use CNN-based solutions to deal with deepfake detection \cite{R6,R20,R24,R25,R26,R27,R34,R48}. These CNN-based methods process input data sequentially, while transformers can process it efficiently by processing the entire input sequence in parallel. Furthermore, in natural language processing (NLP) and computer vision, many transformer-based foundation models trained with large amounts of data have proven useful for downstream application tasks, such as the video vision transformer (ViViT)~\cite{R40}, the audio spectrogram transformer (AST)~\cite{R41}, and the audio-visual transformer (AV-HuBERT)~\cite{R42}. %Inspired by 
Due to the effectiveness of transformers %success 
in various fields \cite{R78,R79,R80}, we intend to apply them to the video forgery detection task by leveraging acoustic and visual features. We believe that multiple modalities provide complementary information and stable inferences \cite{R75,R76,R77}. Therefore, to fill the above gap, we propose a novel \textbf{A}udio-\textbf{V}isual \textbf{T}ransformer-based \textbf{E}nsemble \textbf{Net}work, \textbf{AVTENet}, that utilizes both acoustic and visual features for effective video forgery detection.

Our innovation lies in the application and integration of cutting-edge technologies that were not originally proposed for %the task of 
audio-visual deepfake detection. The main contributions of our work are summarized as follows:

\begin{itemize}
    \item  
    We propose an audio-visual transformer-based ensemble network (AVTENet), and demonstrate that its audio-only, video-only, and audio-visual networks respectively based on cutting-edge pre-trained transformer-based foundation models can be effectively combined to improve robustness and detection accuracy. The proposed approach can address scenarios where only a single modality or both modalities in audio vision are manipulated. 
  
    \item To evaluate the strengths and weaknesses of various ensemble approaches, we present four variants of AVTENet and show that the feature fusion approach performs better than majority voting, score fusion, and average score fusion.
    
    \item We experimentally confirm the effectiveness of AVTENet and empirically demonstrate that our transformer-based ensemble model outperforms the previous CNN-based ensemble network \cite{R6}. 
    
    \item To evaluate the proposed method, we use the FakeAVCeleb dataset \cite{R37}, which is a multimodal audio-visual deepfake dataset. We conduct a comprehensive experimental comparison between the proposed method and existing unimodal and multimodal methods. The results confirm that our proposed method achieves the highest performance on the test sets of the FakeAVCeleb dataset, outperforming all compared methods.

    \item We compare AVTENet's detection capabilities with human detection capabilities.
    
\end{itemize}

The remainder of this paper is organized as follows. Sec. II reviews related work. Sec. III illustrates the proposed methodology. Sec. IV outlines the experimental setup and reports the results. Sec. V discusses the limitations of this study. Finally, Sec VI provides conclusions and future work.

\section{Related Work}
\subsection{Video Deepfake Generation}
Deepfake technology has become more sophisticated as it uses AI techniques to generate, manipulate, or alter video content, thereby adding significant threats to society, legislation, individuals, community consensus, journalism, and cyber-security. Face2Face~\cite{R10}, Neural Talking-Heads~\cite{R8}, First Order Motion Model for Image Animation~\cite{R11}, and Deep Video Portraits~\cite{R12} are some of the deepfake generation techniques. Likewise, there exist different types of deepfake videos, including lip-syncing, face-swapping, and voice manipulation. GANs~\cite{R2}, \cite{R13} have become a popular deep-learning method for generating realistic images or videos. For instance, in \cite{R8} and \cite{R7} GAN-based video deepfake generation methods were proposed. The method proposed in~\cite{R7} generates a dance video of a person from a single image. First, a sequence of human poses is generated and rendered in the video. To generate realistic animated dance videos, this method combines pose estimation, key-point detection, and image synthesis. Similarly, the few-shot learning method proposed in~\cite{R8} uses a few reference images to generate realistic videos of talking heads. Their method creates a realistic video sequence of a target person speaking by mapping several reference images. Another GAN-based method for image-to-image translation was proposed in~\cite{R9}. To ensure high perceptual qualities of the generated images, the model was trained with a perceptual loss function. The authors evaluated their method on several image generation tasks, such as image colorization, painting, and super-resolution.

%\vspace{-2mm}

\subsection{Video Deepfake Detection}
\subsubsection{Unimodal Deepfake Detection}

Previous work has mainly focused on a single modality to detect forgeries in videos since synthetic videos generally have less manipulation in the audio track (Fig. \ref{newFigure}(b)). Some researchers have worked on detecting forgeries in videos by exploiting visual artifacts such as inconsistent facial expressions and movements \cite{R16,R18}, irregular head movements or angles \cite{R23,R16}, mismatched shadows or lightening \cite{R54}, distorted or blurred edges around the face or body \cite{R55}, unnatural eye blinking or movement \cite{R19}, unnatural background or scenery \cite{R56}. The authors of \cite{R22} argued that several facial editing techniques exhibit visual artifacts in videos that can be tracked and used to detect video forgeries. Considering this rationale, they demonstrated that even simple visual cues can reveal deepfakes.
The authors of \cite{R19} used eye blinking, a physiological signal that is not well presented in synthetic videos, to detect fake face videos. Other researchers have exploited emotional inconsistency \cite{R16}, synchronization between speech and lip movements \cite{R20}, or motion analysis \cite{R57}. A biometrics-based forensics technique was proposed in~\cite{R17} to detect faceswap deepfakes by exploiting appearance and behavior, especially facial expressions and head movements. MesoNet and MesoInception are two popular models proposed in~\cite{R24}. Both are compact CNN-based models with a few layers that exploit mesoscopic properties to detect facial tampering in videos. MesoNet analyzes the residual high-frequency content of video frames. XceptionNet \cite{R25} is built on depth-wise separable convolutional layers.  
The CNN-based method proposed in \cite{R26} employs composite coefficients that effectively scale all dimensions of depth, width, and resolution uniformly. 

A transformer network based on an incremental learning strategy is proposed in \cite{R28}, which utilizes face images and their UV texture generated by a 3D face construction method using a single-face input image to detect synthetic videos. Later, the same authors proposed a hybrid transformer network to detect forged videos \cite{R29}. The transformer network is trained end-to-end using XceptionNet and EfficientNet-B4 as feature extractors. Obscene video detection \cite{R74} can be considered related to the task of fake video detection, where most obscene videos of celebrities are fake.

\subsubsection{Multimodal Deepfake Detection}

Considering that the auditory context is crucial in visual object perception \cite{R35}, few recent attempts have addressed the challenge of detecting multimodal deepfake videos using multiple modalities (Fig. \ref{newFigure}(c)) such as video, audio, and text. 
The CNN-based method proposed in \cite{R27} utilizes audio and video streams to extract features and then combines these features to better detect deepfake videos.
A learning-based approach, inspired by Siamese networks and triplet loss, is proposed in \cite{R30}. This approach enhances information for learning by using the acoustic and visual features of the same video. The authors extracted affective cues and used them to compare the emotions perceived in the two modes to discern whether the video was genuine or manipulated. The authors of \cite{R31},  \cite{R32}, and  \cite{R60} combined audio and visual modalities to develop a video deepfake detector that can handle audio-only, video-only, and audio-visual manipulations. In \cite{R31}, a promising approach was proposed to exploit the intrinsic synchronization between audio and visual modalities to potentially detect increasingly prevalent deepfakes. In \cite{R32}, an audio-visual person-of-interest deepfake detector that exploits specific biometrics of 
individual was proposed to handle various manipulation methods used in synthetic media generation. The high generalizability of the detector is due to its independence from any manipulation techniques and training on videos of real talking faces. Similarly,  for audio-visual forgery detection, the authors of \cite{R60} incorporate audio loss, video loss and audio-visual loss to capture the inconsistency of multimodality and artifacts from individual modalities. Additionally, ensemble approaches were used in \cite{R6} and \cite{R33} to improve the accuracy of deepfake detection. The ensemble method in \cite{R6} combines CNN-based audio-only, video-only, and audio-visual networks to detect multimodal deepfakes. In \cite{R33}, incorporating supplementary textual features was demonstrated to be effective in handling multimodal fake content detection. The authors built an ensemble network combining unimodal and crossmodal classifiers to distinguish forged from genuine video clips. %On the other hand
Furthermore, the authors of \cite{R34} argued that the manipulation of audio or visual flow in video can lead to a lack of harmony between them, such as speech-lip non-synchrony and unnatural facial movements. Therefore, they utilized modality dissonance scores to measure the dissimilarity between two modalities to identify genuine or spoofed videos. Another deep learning method that exploits the audio and visual modalities and considers the importance of speech-lip synchronization for the task of video forgery detection was proposed in \cite{R20}. Their audio-visual deepfake detector checks the synchrony between the lip sequence extracted from the video and the synthetic lip sequence generated from the audio track. Similarly, the authors of \cite{R58} adopted a self-supervised transformer-based approach that utilizes contrastive learning. Their method allows paired acoustic and visual streams to learn mouth motion representations propelling paired representation to be closer and unpaired to stay farther %in order 
to determine audio-visual forgery. Another self-supervised learning approach in \cite{R47} exploits an audio-visual temporal synchronization by evaluating consistency between the acoustic stream and faces in a video clip to determine forgery. 
In addition to the above studies, there are also multimodal methods for other tasks worth referring to. For %example
instance, \cite{R65} proposed a two-stage cascade framework based on multimodal data fusion for face anti-spoofing. The authors of \cite{R64} proposed a deep %convolution neural network 
CNN-- called LieNet to detect deception by combining contact and non-contact signal modalities. The authors of \cite{R66} proposed a multimodal feature-fusion-based model called MFFFLD for fingerprint liveness detection. \cite{R67} compared the recognition performance and robustness of different multimodal emotion recognition models.
Although multimodal approaches have been studied in video forgery detection; more exploration is %clearly 
needed.

\begin{figure*}[!t]
\centering
\includegraphics[width=6.0in]{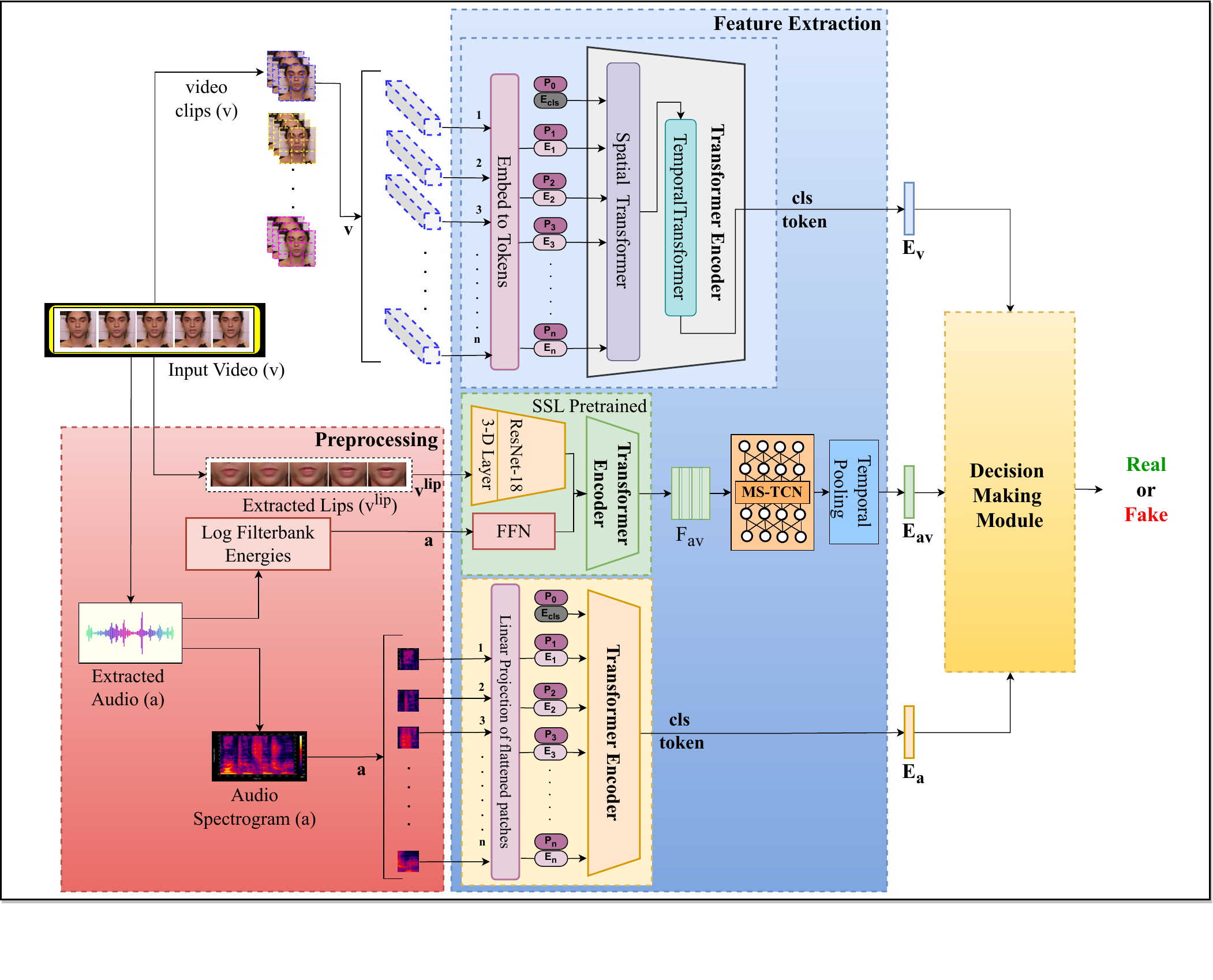}
\vspace{-10mm}
\caption{Overview of the proposed AVTENet model for detecting video forgeries by simultaneously utilizing acoustic and visual cues. It consists of a video-only network (VN), an audio-only network (AN), and an audio-visual network (AVN), each of which is an independent transformer-based classifier, and a decision-making module (DM) that produces final predictions based on the three classifiers through various ensemble strategies. }
\label{Fig2}
\vspace{-4mm}
\end{figure*}

\subsection{Multimodal Deepfake Datasets}\label{sec_datasets} 

To the best of our knowledge, DFDC \cite{R36} and FakeAVCeleb \cite{R37} are the only multimodal datasets that include both acoustic and visual manipulations in videos. 

The DFDC dataset \cite{R36} is larger than the FakeAVCeleb dataset, containing 100,000 video clips collected from 3,426 paid subjects. DFDC provides a benchmark for evaluating the performance of deepfake detection models and serves as a crucial resource for advancing research in %the field of 
deepfake detection. It has been released as part of a competition to promote the development of deepfake detection methods. The deepfake videos in this dataset were generated using various deepfake techniques, such as GAN-based methods and nonlearning methods. It %consists of 
comprises real and fake videos and considers both acoustic and visual manipulations.

The FakeAVCeleb dataset \cite{R37} is a multimodal audio-visual dataset released after DFDC. The dataset was carefully generated through multiple manipulation methods, %taking into account 
considering the balance among ethnic backgrounds, gender, and age groups. The real videos of 500 celebrities originating from the Voxceleb2 dataset \cite{R61} collected from YouTube form the base set. This base set was further used to generate 19,500 deepfake videos through manipulation techniques, including Faceswap \cite{R43}, FSGAN \cite{R44}, Wav2Lip \cite{R45}, and RTVC (real-time-voice-cloning) \cite{R46}.

\section{Proposed Method}
The natural auditory context provides distinctive, independent, and diagnostic information about the visual world, which directly impacts the perceptual experience of visual objects \cite{R73}\cite{R35}. To combat the emerging threat of video deepfakes, inspired from human cognition of multisensory information, we leverage both acoustic and visual information to detect forgeries in videos. We use the transformer-based model because its self-attention mechanism can catch inconsistencies or manipulations to identify video forgeries. Furthermore, we employ ensemble learning to exploit the fusion of complementary information, modality-specific patterns, and multimodal features to improve the robustness of the detection system \cite{R77}. Accordingly, we propose an %audio-visual transformer-based ensemble network 
AVTENet that fully exploits acoustic and visual information to detect video forgeries by leveraging modality-specific cues and complementary insights of joint audio and visual information. Fig.~\ref{Fig2} shows an overview of our proposed AVTENet model, which consists of three key networks, namely a video-only network (VN), an audio-only network (AN), and an audio-visual network (AVN), as well as a decision-making module (DM). The three key networks are different-modality transformer-based networks integrated with pre-trained models through self-supervised learning (SSL) or supervised learning (SL). %Then
Subsequently, DM integrates the outputs of these three classifiers according to different fusion strategies.
Given a test video $x$, the decision is made according to
\begin{equation}\label{eq_AVTENet}
{{\mathrm{AVTENet}(x)} = {\mathrm{DM}(C_{v}(x)}, {C_{a}(x)}, {C_{av}(x))}},
\end{equation}
where $C_v$, $C_a$, and $C_{av}$ denote the video-only, audio-only, and audio-visual classifiers, respectively, and $\mathrm{DM}$ denotes the function of decision making.

The AVTENet architecture is %motivated by its ability
designed to effectively capture and utilize complementary information from both modalities. %The purpose of using 
Three networks are used in our ensemble approach to leverage the strength of each network. Individual unimodal networks (AN and VN) specialize in processing information from specific modalities. AN specializes in analyzing audio data to extract valuable acoustic features that help detect acoustic manipulation, while VN focuses on processing video data to extract visual features that can reveal visual forgeries. Combining them into AVN allows for a more comprehensive analysis of the input and can capture complex relationships and dependencies between the two modalities. These multimodal features provide more reliable detection results and enhance robustness.

\vspace{-2mm}
\subsection{Video-only Network (VN)} 
The VN module extracts relevant spatiotemporal features from video frames through its self-attention mechanism to capture spatiotemporal patterns and long-range dependencies. It then outputs a vector representation that captures essential visual information. 
    
%Inspired by 
Building on the remarkable progress achieved by the video vision transformer ViViT \cite{R40} in video classification tasks, we seamlessly integrate it into the visual backbone of AVTENet. As shown %at the top of 
in Fig.~\ref{Fig2}, we use a factorized encoder model that consists of two separate transformer encoders. The first is a spatial encoder, which delves into the interaction among tokens extracted from the same temporal index. The second is a temporal encoder that models interactions among tokens from different temporal indices. We first divide the input video into small segments, a series of patches is extracted from each segment 
%in the form of 
as a tubelet and passed through the spatial encoder to produce an encoding vector representing the segment. The output encoding vectors serve as input to the temporal transformer along with an additional classification (cls) token to extract the final representation of the entire video. 

For VN training, a dataset $D^v=\{v_i,y_i\}_{i=1}^{n}$ is extracted from the training set of the FakeAVCeleb dataset, where $v_i$ denotes the video stream of the $i$-th training sample $x_i$, and $y_i$ is the label for $v_i$ (0 for fake and 1 for real). As shown in Fig.~\ref{Fig2}, the ViViT model extracts the single-vector representation $E_v$ (the $cls$ token of the temporal transformer) of each training video clip. A linear layer is used as the classifier. The ViViT model is pre-trained on the Kinetics dataset ~\cite{R68}. During the VN training process, %not only is 
the linear classification layer is trained, and %but also 
the pre-trained ViViT model is fine-tuned. The model is trained with the binary cross-entropy loss defined as
\begin{equation}
{\mathcal{L}_\mathrm{BCE}=-\frac{1}{n} \sum_{i=1}^{n} y_{i} \log (\hat{y}_{i}) + \left(1-y_{i}\right) \log \left(1-\hat{y}_{i}\right)},
\label{eq_5}
\end{equation}
where $n$ is the total number of training samples, and $\hat{y}_{i}$ denotes the predicted score.

In the inference phase of VN, given a test sample $x$, the prediction is conducted by
\begin{equation}\label{eq_VN}
C_v(x) = \mathrm{VN}(x_v),
\end{equation}
where $x_v$ is the video stream (i.e., the image sequence) of $x$. %Note that 
In the inference phase of our ensemble model, the DM module utilizes either the single-vector representation or the predicted score corresponding to the ``fake'' of the video-only network. 

\vspace{-2mm}
\subsection{Audio-only Network (AN)}
The AN module takes as input an audio spectrogram, where time and frequency elements help learn and capture acoustic patterns, temporal dynamics, and other audio-specific characteristics, and returns a high-dimensional feature representation. 

Transformers perform incredibly well in audio processing as their self-attention mechanisms allow for capturing long-range dependencies in audio. In this %work
study, we use the audio spectrogram transformer (AST) \cite{R41} as the audio backbone of AVTENet. As shown %at the bottom of 
in Fig. \ref{Fig2}, the AN module takes as input a spectrogram that is further divided into a sequence of 16x16 patches with an overlap of six in both the time and frequency dimensions. A linear projection layer transforms each patch into a 1D embedding. Since the patch sequence lacks input order information, trainable positional embeddings are incorporated into each patch embedding to enable capturing spatiotemporal structure from a 2D audio spectrogram. Additionally, a classification (cls) token is added to the sequence. The output cls embedding of the transformer encoder encapsulates information about characteristics that can be used to determine challenging and deceptive alterations in the audio.  

For training AN, a dataset $D^a=\{a_i,y_i\}_{i=1}^{n}$ is extracted from the training set of the FakeAVCeleb dataset, where $a_i$ denotes the audio track of the $i$-th training sample $x_i$, and $y_i$ is the label for $a_i$ (0 for fake and 1 for real). As shown in Fig. \ref{Fig2}, the AST model is used to extract the single vector representation $E_a$ (the $cls$ token) of the mel-spectrogram of each training audio clip. A linear layer is used as the classifier. The AST model is pre-trained on the Audioset \cite{R41}. During the training of AN, the linear classification layer is trained, and the pre-trained AST model is fine-tuned. As with the video-only network, the cross-entropy loss is used to train the audio-only network. 

In the inference phase of AN, given a test sample $x$, the prediction is conducted by
\begin{equation}\label{eq_AN}
C_a(x) = \mathrm{AN}(x_a),
\end{equation}
where $x_a$ is the audio track of $x$.
As with the video-only network, in the inference phase of our ensemble model, either the single-vector representation or the predicted score corresponding to the ``fake'' of the audio-only network is used by the DM module.

\begin{figure}[!b]
\vspace{-6mm}
\centering
\includegraphics[width=3.0in]{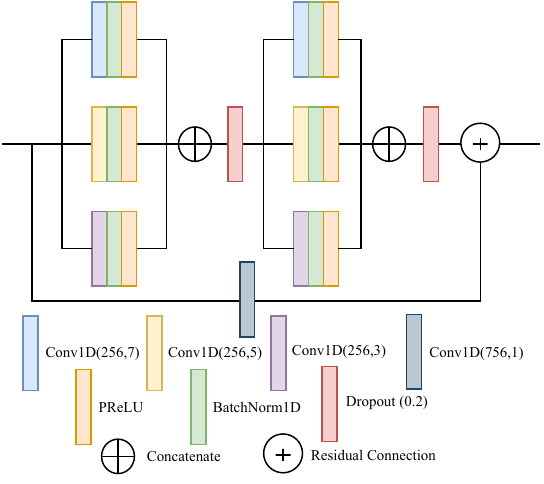}
\vspace{-4mm}
\caption{Architecture of MS-TCN block.}
\label{Fig}
% \vspace{-8mm}
\end{figure}

\subsection{Audio-Visual Network (AVN)} 
The AVN module exploits both visual and acoustic information by jointly learning paired audio and visual inputs that provide complementary information and capture meaningful representations between these distinct modalities to expose fabrication in videos. 

We use the self-supervised learning framework AV-HuBERT~\cite{R42} as the audio-visual backbone of AVTENet. As shown in %the middle of
Fig.~\ref{Fig2}, the AVN module simultaneously uses the log filterbank energies and the video frame stack of the lip region extracted from the video as inputs to the lightweight modality-specific encoders. The frames of the lip region are passed through a ResNet-based feature extractor to extract relevant visual features. %At the same time
Simultaneously, the log filterbank energy features are passed through the feed-forward network (FFN) to extract the acoustic features. The visual and acoustic features are then fused and fed to a shared transformer encoder to extract jointly contextualized audio-visual representations that encapsulate the correlation between acoustic and visual modalities. These audio-visual representations are further passed through a multiscale temporal convolutional network (MS-TCN) and a temporal pooling layer to produce a single-vector representation. Fig.~\ref{Fig} shows the structural diagram of MS-TCN. The details can be found in \cite{R53} and \cite{R69}.

\begin{figure}[!t]
%\vspace{-4mm}
\centering
\includegraphics[width=3.5in]{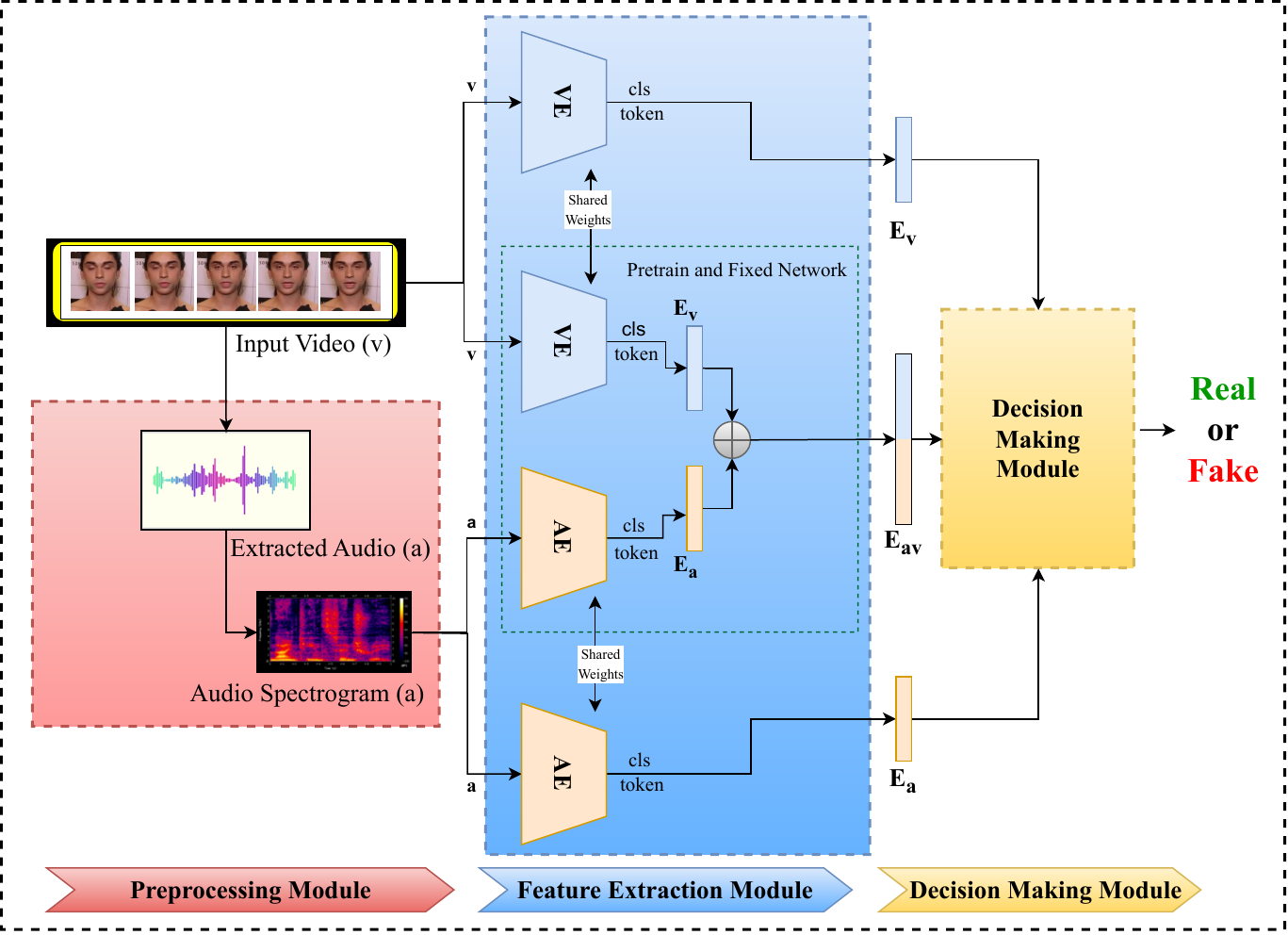}
\vspace{-6mm}
\caption{Overview of another AVTENet model that consists of AST-based AN, ViViT-based VN, and AST\&ViViT-based AVN.}
\label{Fig3}
\vspace{-6mm}
\end{figure}

For training AVN, a dataset $D^{av}=\{a_i,v_i^{lip}, y_i\}_{i=1}^{n}$ is extracted from the training set of the FakeAVCeleb dataset, where $a_i$ and $v_i^{lip}$, respectively, denote the audio track and the lip image sequence of the $i$-th training sample $x_i$, and $y_i$ is the label for $x_i$ (0 for fake and 1 for real). As shown in Fig.~\ref{Fig2}, the AV-HuBERT model, MS-TCN network, and temporal pooling layer are used to extract the vector representation $E_{av}$ of the log filterbank energy features of the audio track and the lip-image sequence of each training video clip. A linear layer is used as the classifier. The AV-HuBERT model is pre-trained on the LRS3 dataset~\cite{R62}. During the training process of AVN, both the MS-TCN with a linear classification layer and the pre-trained AV-HuBERT model are trained, with the latter being fine-tuned.

\begin{figure*}[!t]
% \vspace{-4mm}
\centering
\vspace{-6mm}
\includegraphics[width=7.4in]{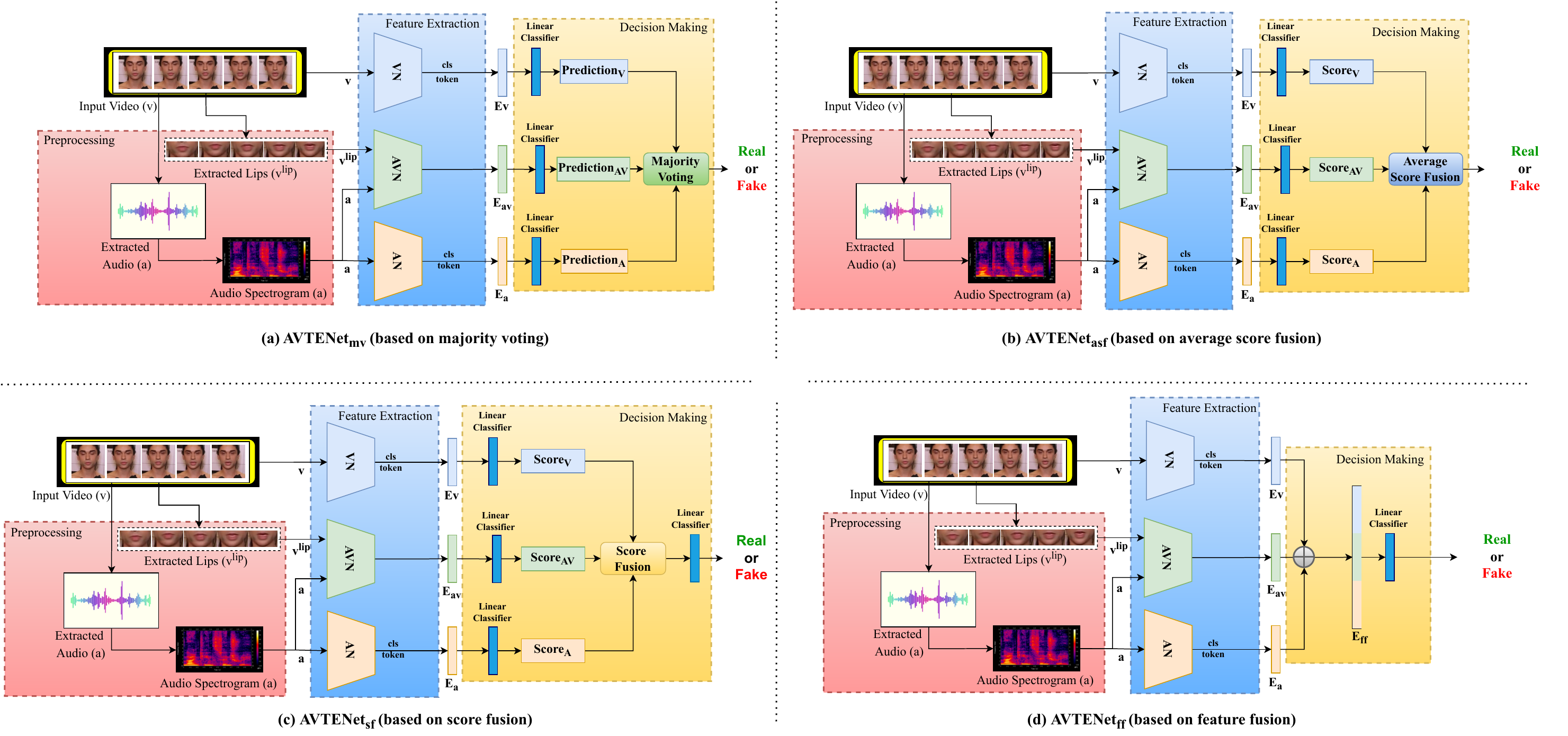}
\vspace{-4mm}
\caption{Four variants of the proposed AVTENet model for video forgery detection based on different ensemble strategies: (a) $AVTENet_{mv}$ based on majority voting, (b) $AVTENet_{asf}$ based on average score fusion, (c) $AVTENet_{sf}$ based on score fusion, and (d) $AVTENet_{ff}$ based on feature fusion.}
\label{Fig4}
\vspace{-4mm}
\end{figure*}

In the inference phase of AVN, given a test sample $x$, the prediction is conducted by
\begin{equation}\label{eq_AVN}
C_{av}(x) = \mathrm{AVN}(x_a,x_{v^{lip}}),
\end{equation}
where $x_a$ and $x_{v^{lip}}$ are the audio track and the lip-image sequence of $x$, respectively.
As with the video-only and audio-only networks, in the inference phase of our ensemble model, either the single-vector representation or the predicted score corresponding to the ``fake'' of the audio-visual network is used by the DM module.

In addition to the above AV-HuBERT-based AVN setup, we also study another setup. The audio and video embeddings are extracted via AST (same as AN) and ViViT (same as VN). The corresponding $cls$ tokens are concatenated as $E_{av}$. The prediction is also conducted by Eq. (5), but using ($x_a$, $x_v$) rather than ($x_a$, $x_{v^{lip}}$) as input. An overview of the AVTENet model based on this AVN setting is shown in  Fig. \ref{Fig3}. Compared with the AVTENet model in Fig. \ref{Fig2}, the only difference is that the AV-HuBERT-based AVN is replaced by the AST\&ViViT-based AVN.

\subsection{Decision-Making Module (DM):}
As shown in Fig. \ref{Fig2}, the DM module of AVTENet is used to integrate the outputs of VN, AN, and AVN (trained separately as described previously), to produce final predictions that indicate potential forgery in the video. In this study, we investigate several fusion strategies.        
  
\subsubsection{Majority Voting}
As shown in Fig. \ref{Fig4}(a) and Algorithm \ref{alg1}, the DM module outputs ``fake'' if at least two of VN, AN, and AVN output ``fake''. The corresponding AVTENet model is termed $\mathrm{AVTENet}_{mv}$. No %Note that no
additional models need to be trained for majority voting fusion. 

\subsubsection{Average Score Fusion}
As shown in Fig. \ref{Fig4}(b), the DM module outputs ``fake'' if the average of the output scores of VN, AN, and AVN exceeds a preset threshold. The corresponding AVTENet model is termed $\mathrm{AVTENet}_{asf}$. No %Note that no 
additional models need to be trained for average score fusion.

\subsubsection{Score Fusion}
As shown in Fig.~\ref{Fig4}(c), in score fusion, the output scores of VN, AN, and AVN are concatenated as input to a linear layer trained on the same training data as the three component classifiers with the cross-entropy loss. When training the linear layer, the parameters of VN, AN, and AVN are fixed. The corresponding AVTENet model is termed $\mathrm{AVTENet}_{sf}$.

\subsubsection{Feature Fusion}
As shown in Fig.~\ref{Fig4}(d), in feature fusion, the output representations of the penultimate layers of VN, AN, and AVN are concatenated as input to a linear layer trained on the same training data as the three component classifiers with the cross-entropy loss. When training the linear layer, the parameters of VN, AN, and AVN are fixed. The corresponding AVTENet model is termed $\mathrm{AVTENet}_{ff}$.

\begin{algorithm}[!b]
\caption{Majority Voting Algorithm for $\mathrm{AVTENet}_{mv}$}
\label{alg1}
\begin{algorithmic}[1]
\REQUIRE $P_v, P_a$, and $P_{av} \left(P_i \in 0\right.$ or 1$)$
\ENSURE AVTENet\_Prediction $P  \left(P \in 0\right.$ or 1$)$
\STATE Initialize $vote\_count$ variable, $vote\_count$ = $P_v + P_a + P_{av}$
    \IF{ $vote\_count$ $\geq 2$}
        \STATE $P$ = 1
    \ELSE
              \STATE $P$ = 0
    \ENDIF
\RETURN $P$
\end{algorithmic}
\end{algorithm}

\section{Experimental Results}
This section presents the experiment setup, including the dataset, data analysis and preprocessing, evaluation metrics, training hyperparameters, and experimental results.  

\vspace{-2mm}
\subsection{Dataset}

\begin{figure}[!t]

\centering
\includegraphics[width=3.3in]{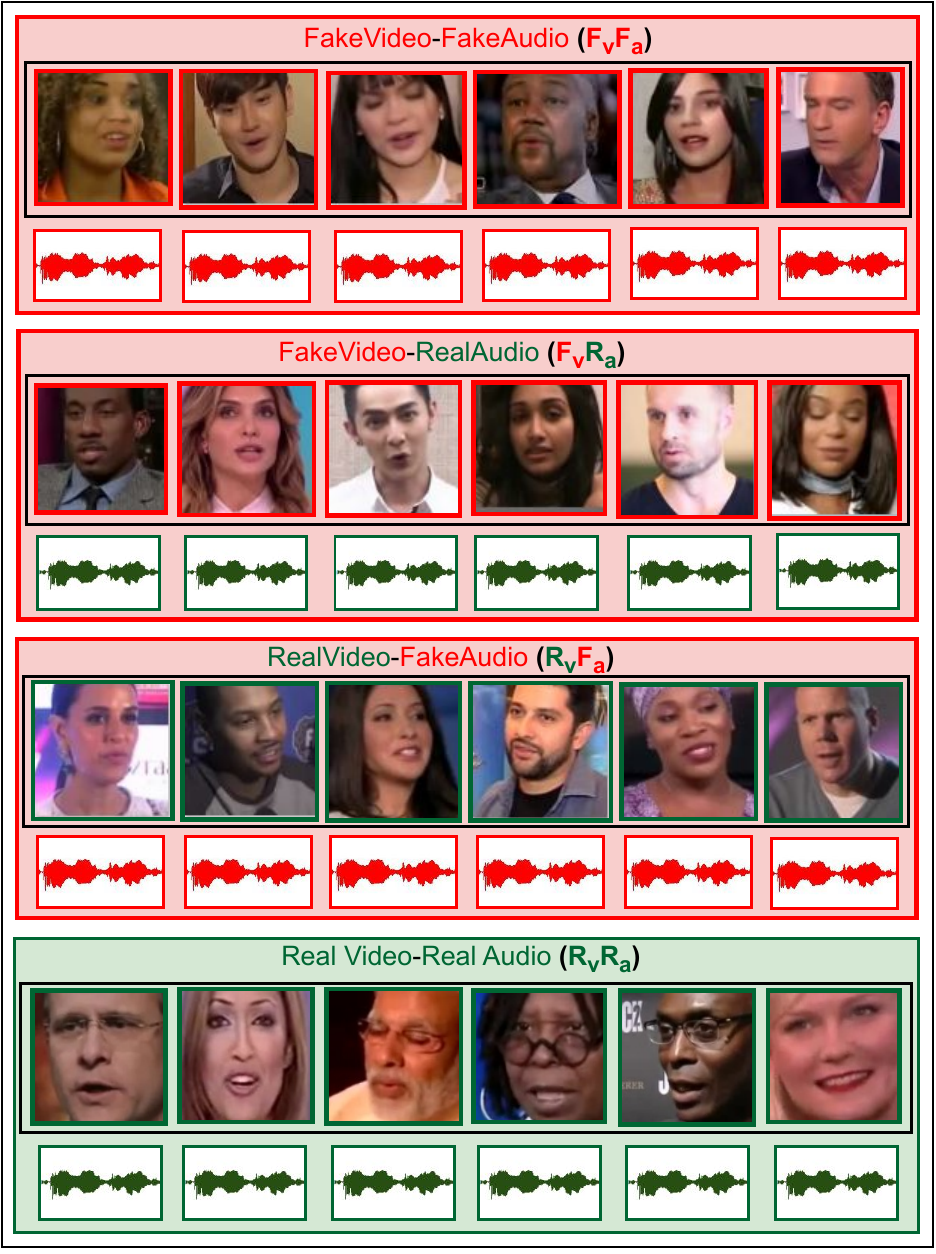}
\vspace{-2mm}
\caption{Illustrations of the audio and visual streams of some samples selected from the FakeAVCeleb Dataset, where real streams are indicated in green, and fake ones are indicated in red.}
\label{Fig5}
\vspace{-4mm}
\end{figure}

Our experiments are conducted on the FakeAVCeleb dataset\footnote{We do not use the DFDC dataset for two reasons. First, the subject's face in some videos may not be facing the camera, and our model needs to extract lip images. Second, its labeling lacks separate labels for acoustic and visual modalities in videos.} \cite{R37}. As described in Sec. \ref{sec_datasets}, it contains 500 videos featuring 500 celebrities and 19,500 fake videos manipulated from these 500 videos. The videos corresponding to 70 celebrities are used for testing, and the videos corresponding to the remaining 430 celebrities are used for training. These videos are divided into four categories according to whether the audio and video modalities are manipulated, namely RealVideo-RealAudio ($R_{v}R_{a}$), RealVideo-FakeAudio ($R_{v}F_{a}$), FakeVideo-RealAudio ($F_{v}R_{a}$) and FakeVideo-FakeAudio ($F_{v}F_{a}$), where each video is carefully labeled to indicate its authenticity and facilitate training and testing forgery detection models. Fig.~\ref{Fig5} shows a few samples of each category from the FakeAVCeleb dataset.

\subsubsection{Testing Sets}
To conduct a comprehensive evaluation of various unimodal (AN and VN) and multimodal (AVN and AVTENet) detection methods using the FakeAVCeleb dataset, we used eight different test sets, including Testset-I, Testset-II, faceswap, faceswap-wav2lip, fsgan, fsgan-wav2lip, RTVC, and wav2lip. Except for Testset-I and Testset-II, each test set contains the same set of 70 genuine videos of 70 subjects unseen in training and a different set of 70 fake videos generated using specific manipulation techniques. Testset-I and Testset-II are the main evaluation test sets, which also contain the above 70 genuine videos. However, for the 70 fake videos, Testset-I contains the same number of fake samples from each manipulation technique, while Testset-II contains an equal number of fake samples from the $R_{v}F_{a}$, $F_{v}F_{a}$ and $F_{v}R_{a}$ categories. Table~\ref{table2} summarizes the manipulation modalities involved in each test set. Multiple test sets can provide an exhaustive understanding of the strengths and limitations of AVTENet and individual classifiers on fake video samples generated using different deepfake techniques.    

\begin{table}[!h]
\vspace{-2mm}
\caption{Manipulation modalities in each test set. $\checkmark$ indicates manipulation, while $X$ indicates no manipulation.}
\begin{center}
\scalebox{0.8}{
\resizebox{\columnwidth}{!}{%
\begin{tabular}{|c|c|c|}
\hline \textbf{Testsets} & \textbf{Audio Manipulation} & \textbf{Video Manipulation} \\
\hline
\hline Testset-1 & $\checkmark$ & $\checkmark$ \\
\hline Testset-II & $\checkmark$ & $\checkmark$ \\
\hline faceswap & $X$ & $\checkmark$ \\
\hline faceswap-wav2lip & $\checkmark$ & $\checkmark$ \\
\hline fsgan & $X$ & $\checkmark$ \\
\hline fsgan-wav2lip & $\checkmark$ & $\checkmark$ \\
\hline RTVC & $\checkmark$ & $X$ \\
\hline wav2lip & $\checkmark$ & $\checkmark$ \\
\hline
\end{tabular}}
}
\label{table2}
\end{center}
\vspace{-4mm}
\end{table}

\subsubsection{Training Sets for Different Models}
In the experiments, the training data are video samples corresponding to 430 subjects in FakeAVCeleb. However, training unimodal and multimodal classifiers requires specific data settings, so we customize the dataset according to the requirements of each network. Table~\ref{table4} shows the training data setting for each network according to the categorical labels of FakeAVCeleb. Table~\ref{table5} shows the statistics of the real and fake samples used to train each network.

\begin{table}[!h]
% \vspace{-4mm}
\caption{Training data settings for different networks.}
\begin{center}
\scalebox{0.6}{
\resizebox{\columnwidth}{!}{%
\begin{tabular}{|c|c|c|}
\hline
\textbf{Classifier} & \textbf{Class} & \textbf{Category} \\
\hline \hline
\multirow{2}{*}{VN} & Fake & $F_{v}F_{a}$, $F_{v}R_{a}$\\ 
\cline{2-3}
& Real & $R_{v}F_{a}$, $R_{v}R_{a}$ \\ 
\cline{2-3}
\hline
\multirow{2}{*}{AN} & Fake & $R_{v}F_{a}$, $F_{v}F_{a}$\\ 
\cline{2-3}
& Real & $ F_{v}R_{a}$, $ R_{v}R_{a}$ \\ 
\cline{2-3}
\hline
\multirow{2}{*}{AVN} & Fake & $ F_{v}F_{a}$, $F_{v}R_{a}$, $R_{v}F_{a}$\\ 
\cline{2-3}
& Real &$ R_{v}R_{a}$ \\ 
\cline{2-3}
\hline
\end{tabular}}
}
\label{table4}
\end{center}
\vspace{-7mm}
\end{table}

\noindent\textbf{Video-only Network (VN):} The VN classifier is trained to exploit manipulations in visual content. When constructing the training set for VN, as shown in Table \ref{table4}, the fake class includes training samples in the $F_{v}F_{a}$ and $F_{v}R_{a}$ categories because the visual content in these categories is manipulated. In contrast, the real class includes samples from the $R_{v}F_{a}$ and $R_{v}R_{a}$ categories, whose visual content is not manipulated. As shown in Table \ref{table5}, this setting results in the VN training set containing 17,809 video samples per class, of which the fake class contains 9,411 samples from $F_{v}F_{a}$ and 8,398 samples from $F_{v}R_{a}$, while the real class contains 430 samples from $R_{v}R_{a}$, 430 samples from $R_{v}F_{a}$, and 16,949 samples from the external VoxCeleb1 dataset \cite{R38}. External data is used to balance the training data across classes.  

\begin{table}[!h]
\vspace{-2mm}
\caption{Number of training samples for different networks.}
\begin{center}
\scalebox{0.7}{
\resizebox{\columnwidth}{!}{%
\begin{tabular}{|c|c|c|c|}
\hline
\textbf{Classifier} & \textbf{Class} & \textbf{Samples}  & \textbf{Total Training Samples} \\
\hline \hline
\multirow{2}{*}{VN} & {Fake} & 17,809 & \multirow{2}{*}{35,618}\\ 
\cline{2-3}
& {Real} & 17,809 &\\ 
\hline
\multirow{2}{*}{AN } & {Fake} & 9,841 & \multirow{2}{*}{18,669} \\
\cline{2-3}
& {Real} & 8,828 &\\
\hline
\multirow{2}{*}{AVN} & {Fake} & 18,239 & \multirow{2}{*}{36,411}\\
\cline{2-3}
& {Real} & 18,172 &\\
\hline
\end{tabular}}
}
\label{table5}
\end{center}
\vspace{-4mm}
\end{table}

\noindent\textbf{Audio-only Network (AN):} AN learns to detect acoustic manipulations in videos. Therefore, the training set for the AN classifier is organized by treating categories with manipulated acoustic streams as fake and categories with unmanipulated acoustic streams as real. As shown in Table \ref{table4}, the fake class contains all samples of $R_{v}F_{a}$ and $F_{v}F_{a}$, while the real class contains samples from $F_{v}R_{a}$ and $R_{v}R_{a}$. As shown in Table \ref{table5}, the training set for AN contains 9,841 samples in the fake class and 8,828 samples in the real class. For the fake class, 430 samples belong to $R_{v}F_{a}$ and 9,411 samples belong to $F_{v}F_{a}$, while for the real class, 8,398 samples belong to $F_{v}R_{a}$ and 430 samples belong to $R_{v}R_{a}$.

\noindent\textbf{Audio-Visual Network (AVN):} As shown in Table \ref{table4}, for the AVN classifier, we specify $F_{v}F_{a}$, $F_{v}R_{a}$ and $R_{v}F_{a}$ categories, where acoustic and/or visual streams are manipulated, to the fake class. However, the real class only contains video samples from the $R_{v}R_{a}$ category without acoustic and visual manipulations. As shown in Table \ref{table5}, the AVN training set contains 36,411 samples: 18,239 samples for the fake class and 18,172 for the real class. For the fake class, 9,411 samples belong to $F_{v}F_{a}$, 8,398 samples belong to $F_{v}R_{a}$, and 430 samples belong to $R_{v}F_{a}$. For the real class, in addition to the 430 samples from $R_{v}R_{a}$, we borrow 17,742 samples from the VoxCeleb1 dataset to balance the training data.

\subsubsection{Data Analysis and Preprocessing} 

Each classifier involves different data processing steps to ensure effective training and evaluation. The video-only network takes as input a short clip of video; therefore, we split the video into multiple short clips and extract a sequence of consecutive frames from each clip to feed into the network. To prepare the data for the audio-only network, we extract the audio track with a sampling rate of 16kHz from each video. Furthermore, we extract mel-spectrograms from these audios, which are then used as input to the audio-only network. We also extract log filterbank energy features that are used as one of the inputs to the audio-visual network. For the AV-HuBERT-based AVN, we extract the lip image frames from each video and pair them with the corresponding melspectral features as input to the network. 

\vspace{-2mm}
\subsection{Hyperparameters}
We use the Adam optimizer to train each classifier. AN (based on AST) is trained with a learning rate of 0.00001, VN (based on ViViT) is trained with a learning rate of 0.0001, and AVN (with AV-HuBERT) is trained with a learning rate of 0.002. Similarly, AVN (with AST\&ViViT) is trained with a learning rate of 0.0001. 
In AVTENet, the linear layers for combining AST-based AN, ViViT-based VN, and AVN (both AVHuBERT-based and AST\&ViViT-based) are trained with the learning rate of 0.002.

\vspace{-2mm}
\subsection{Evaluation Metrics}
We use accuracy, precision, recall, and F1-score to evaluate the performance of individual unimodal/multimodal classifiers and our AVTENet classifier. 
They are calculated as follows: 
\begin{equation}\label{eq_accuracy}
{Accuracy} = {\frac{TP+TN}{TP+TN+FP+FN}},
\end{equation}
\begin{equation}\label{eq_precision}
    {Precision} = {\frac{TP}{TP+FP}},
\end{equation}
\begin{equation}\label{eq_recall}
{Recall} = {\frac{TP}{TP+FN}},
\end{equation}
\begin{equation}\label{eq_f1}
{F1-score} = {\frac{2\times Precision\times Recall}{Precision+Recall}},
\end{equation}
where ${TP}$, ${TN}$, ${FP}$, and ${FN}$ denote True Positive (fake videos correctly detected as fake), True Negative (real videos correctly detected as real), False Positive (real videos incorrectly detected as fake), and False Negative (fake videos incorrectly detected as real), respectively.
For all metrics, a higher value indicates better performance. 

\subsection{Experimental Results}
This section provides detailed experimental results and analysis. Testset-I and Test-II are the main evaluation test sets of the FakeAVCeleb dataset. 

\subsubsection{Results of the Video-only Network}

\begin{table}[!t]
% \vspace{-6mm}
\caption{Detection Results of the Video-only Network.}
\centering
\scalebox{0.8}{
\begin{tabular}{|c|c|c|c|c|c|c|}
\hline
\textbf {Test-Set type} & \textbf {Class} & \textbf {Precision} & \textbf {Recall} & \textbf {F1-Score} & \textbf {Accuracy} \\
\hline \hline
\multirow{2}{*}{Testset-I} & Real & 0.87 & 0.87 & 0.87 & \multirow{2}{*}{0.87} \\
\cline{2-5}
& Fake & 0.87 & 0.87 & 0.87 & \\ 
\hline
\multirow{2}{*}{Testset-II} & Real & 0.76 & 0.87 & 0.81 & \multirow{2}{*}{0.80} \\
\cline{2-5}
& Fake & 0.85 & 0.73 & 0.78 & \\ 
\hline
\multirow{2}{*}{faceswap} & Real & 1.00 & 0.87 & 0.93 & \multirow{2}{*}{0.94}\\
\cline{2-5}
& Fake & 0.89 & 1.00 & 0.94  & \\ 
\hline
\multirow{2}{*}{faceswap-wav2lip} & Real & 1.00 & 0.87 & 0.93 & \multirow{2}{*}{0.94}  \\
\cline{2-5}
& Fake & 0.89 & 1.00 & 0.94 & \\ 
\hline
\multirow{2}{*}{fsgan} & Real & 0.95 & 0.87 & 0.91 & \multirow{2}{*}{0.91} \\
\cline{2-5}
& Fake & 0.88 & 0.96 & 0.92 & \\ 
\hline
\multirow{2}{*}{fsgan-wav2lip} & Real & 1.00 & 0.87 & 0.93 & \multirow{2}{*}{0.94} \\
\cline{2-5}
& Fake & 0.89 & 1.00 & 0.94 & \\ 
\hline
\multirow{2}{*}{RTVC} & Real & 0.50 & 0.87 & 0.64 & \multirow{2}{*}{0.50} \\
\cline{2-5}
& Fake & 0.50 & 0.13 & 0.20 & \\ 
\hline
\multirow{2}{*}{wav2lip} & Real & 0.97 & 0.87 & 0.92 & \multirow{2}{*}{0.92} \\
\cline{2-5}
& Fake & 0.88 & 0.97 & 0.93 & \\ 
\hline

\end{tabular}}
\vspace{-6mm}
\label{table6}
\end{table}

The results of the VN classifier on eight test sets are shown in Table~\ref{table6}. We observe that the VN classifier %has better performance 
performs better on the test sets %containing
with visual manipulations because it utilizes visual features to detect forgeries in videos. The accuracy on the faceswap, faceswap-wav2lip, fsgan, fsgan-wav2lip, and wav2lip test sets are all above 0.90, except for RTVC, which achieves an accuracy of 0.50. Table~\ref{table2} %evidently 
shows that, except for RTVC, all these test sets contain visual manipulations, making VN perform better on these test sets. The poor performance on RTVC is due to the lack of visual manipulations. 
Notably, both faceswap-wav2lip and fsgan-wav2lip include acoustic manipulations as well, but the VN classifier only detects visual manipulations in these cases. 
Furthermore, the accuracy of the two main test sets (0.87 for Test set-I and 0.80 for Testset-II) is relatively low compared to other test sets. This is because, for both test sets, the fake class contains videos with or without visual manipulation. In conclusion, %It is concluded that 
the VN classifier performs %well 
better when the video is visually manipulated, but fails %completely 
when the video only contains acoustic manipulation.

\subsubsection{Results of the Audio-only Network}
The AN classifier is a unimodal method that uses acoustic features to detect whether a video is real or fake. The results of AN on eight test sets are shown in Table~\ref{table7}. We observe a similar pattern to the results of the VN classifier in Table \ref{table6}. The detection accuracy is almost perfect on the faceswap-wav2lip, fsgan-wav2lip, and RTVC test sets (all containing acoustic manipulation in the fake class). %But 
However, AN fails %completely 
on the faceswap and fsgan test sets, where fake instances are only visually manipulated while the audio tracks are all real. Furthermore, the detection accuracies of Testset-I and Testset-II (0.80 and 0.84, respectively) are lower than the strong performance of the other test sets. This is because, for both test sets, the fake class contains videos with or without acoustic manipulation. The results %of this 
how that the AN classifier can effectively detect acoustically manipulated videos, but cannot detect fake videos containing only visual manipulations.

\begin{table}[!b]
\vspace{-6mm}
\caption{Detection Results of the Audio-only Network.}
\centering
\scalebox{0.8}{
\begin{tabular}{|c|c|c|c|c|c|c|}
\hline
\textbf {Test-Set type} & \textbf {Class} & \textbf {Precision} & \textbf {Recall} & \textbf {F1-Score} & \textbf {Accuracy} \\
\hline \hline
\multirow{2}{*}{Testset-I} & Real & 0.71 & 1.00 & 0.83 & \multirow{2}{*}{0.80}  \\
\cline{2-5}
& Fake & 1.00 & 0.60 & 0.75 & \\ 
\hline
\multirow{2}{*}{Testset-II} & Real & 0.76 & 1.00 & 0.86 & \multirow{2}{*}{0.84} \\
\cline{2-5}
& Fake & 1.00 & 0.69 & 0.81 & \\ 
\hline
\multirow{2}{*}{faceswap} & Real & 0.50 & 1.00 & 0.67 & \multirow{2}{*}{0.50}  \\
\cline{2-5}
& Fake & 0.00 & 0.00 & 0.00  & \\ 
\hline
\multirow{2}{*}{faceswap-wav2lip} & Real & 1.00 & 1.00 & 1.00 & \multirow{2}{*}{1.00}   \\
\cline{2-5}
& Fake & 1.00 & 1.00 & 1.00 & \\ 
\hline
\multirow{2}{*}{fsgan} & Real & 0.50 & 1.00 & 0.67 & \multirow{2}{*}{0.50} \\
\cline{2-5}
& Fake & 0.00 & 0.00 & 0.00 & \\ 
\hline
\multirow{2}{*}{fsgan-wav2lip} & Real & 1.00 & 1.00 & 1.00 & \multirow{2}{*}{1.00} \\
\cline{2-5}
& Fake & 1.00 & 1.00 & 1.00 & \\ 
\hline
\multirow{2}{*}{RTVC} & Real & 0.97 & 1.00 & 0.99 & \multirow{2}{*}{0.99} \\
\cline{2-5}
& Fake & 1.00 & 0.97 & 0.99 & \\ 
\hline
\multirow{2}{*}{wav2lip} & Real & 0.69 & 1.00 & 0.82 & \multirow{2}{*}{0.78} \\
\cline{2-5}
& Fake & 1.00 & 0.56 & 0.72 & \\ 
\hline
\end{tabular}}
\label{table7}
% \vspace{-6mm}
\end{table}

\begin{table}[!t]
% \vspace{-8mm}
\caption{Detection Results of the AV-HuBERT-based Audio-Visual Network.}
\centering
\scalebox{0.8}{
\begin{tabular}{|c|c|c|c|c|c|c|}
\hline
\textbf {Test-Set type} & \textbf {Class} & \textbf {Precision} & \textbf {Recall} & \textbf {F1-Score} & \textbf {Accuracy} \\
\hline \hline
\multirow{2}{*}{Testset-I} & Real & 0.92 & 0.96 & 0.94 & \multirow{2}{*}{0.94}  \\
\cline{2-5}
& Fake & 0.96 & 0.91 & 0.93 & \\ 
\hline
\multirow{2}{*}{Testset-II} & Real & 1.00 & 0.96 & 0.98 & \multirow{2}{*}{0.98}  \\
\cline{2-5}
& Fake & 0.96 & 1.00 & 0.98 & \\ 
\hline
\multirow{2}{*}{faceswap} & Real & 0.79 & 0.96 & 0.86 & \multirow{2}{*}{0.85}  \\
\cline{2-5}
& Fake & 0.95 & 0.74 & 0.83  & \\ 
\hline
\multirow{2}{*}{faceswap-wav2lip} & Real & 1.00 &  0.96 & 0.98 & \multirow{2}{*}{0.98} \\
\cline{2-5}
& Fake & 0.96 & 1.00 & 0.98 & \\ 
\hline
\multirow{2}{*}{fsgan} & Real & 0.91 & 0.96 & 0.93 & \multirow{2}{*}{0.93} \\
\cline{2-5}
& Fake & 0.95 & 0.90 & 0.93 & \\ 
\hline
\multirow{2}{*}{fsgan-wav2lip} & Real & 1.00 & 0.96 & 0.98 & \multirow{2}{*}{0.98} \\
\cline{2-5}
& Fake & 0.96 & 1.00 & 0.98 & \\ 
\hline
\multirow{2}{*}{RTVC} & Real & 0.96 & 0.96 & 0.96 & \multirow{2}{*}{0.96}  \\
\cline{2-5}
& Fake & 0.96 & 0.96 & 0.96 & \\ 
\hline
\multirow{2}{*}{wav2lip} & Real & 1.00 & 0.96 & 0.98 & \multirow{2}{*}{0.98} \\
\cline{2-5}
& Fake & 0.96 & 1.00 & 0.98 &  \\ 
\hline
\end{tabular}}
\label{table8}
\vspace{-6mm}
\end{table}

\begin{table}[!b]
\vspace{-6mm}
\caption{Detection Results of $AVTENet_{mv}$ (Majority Voting).}
\centering
\scalebox{0.8}{
\begin{tabular}{|c|c|c|c|c|c|c|}
\hline
\textbf {Test-Set type} & \textbf {Class} & \textbf {Precision} & \textbf {Recall} & \textbf {F1-Score} & \textbf {Accuracy} \\
\hline \hline
\multirow{2}{*}{Testset-I} & Real & 0.92 & 1.00 & 0.96 & \multirow{2}{*}{0.96} \\
\cline{2-5}
& Fake & 1.00 & 0.91 & 0.96 & \\ 
\hline
\multirow{2}{*}{Testset-II} & Real & 1.00 & 1.00 & 1.00 & \multirow{2}{*}{1.00}  \\
\cline{2-5}
& Fake & 1.00 & 1.00 & 1.00 & \\ 
\hline
\multirow{2}{*}{faceswap} & Real & 0.80 & 1.00 & 0.89 & \multirow{2}{*}{0.87}  \\
\cline{2-5}
& Fake & 1.00 & 0.74 & 0.85  & \\ 
\hline
\multirow{2}{*}{faceswap-wav2lip} & Real & 1.00 &  1.00 & 1.00 & \multirow{2}{*}{1.00} \\
\cline{2-5}
& Fake & 1.00 & 1.00 & 1.00 & \\ 
\hline
\multirow{2}{*}{fsgan} & Real & 0.88 & 1.00 & 0.93 & \multirow{2}{*}{0.93}  \\
\cline{2-5}
& Fake & 1.00 & 0.86 & 0.92 & \\ 
\hline
\multirow{2}{*}{fsgan-wav2lip} & Real & 1.00 &  1.00 &  1.00 & \multirow{2}{*}{ 1.00}  \\
\cline{2-5}
& Fake &  1.00 & 1.00 &  1.00 & \\ 
\hline
\multirow{2}{*}{RTVC} & Real & 0.96 & 1.00 & 0.98 & \multirow{2}{*}{0.98}  \\
\cline{2-5}
& Fake & 1.00 & 0.96 & 0.98 & \\ 
\hline
\multirow{2}{*}{wav2lip} & Real & 0.99 & 1.00 & 0.99 & \multirow{2}{*}{0.99}  \\
\cline{2-5}
& Fake & 1.00 & 0.99 & 0.99 & \\ 
\hline

\end{tabular}}
\label{table9}
% \vspace{-6mm}
\end{table}
\subsubsection{Results of the Audio-Visual Network}
Compared to the VN and AN classifiers, which can only detect visual and acoustic manipulations in fake videos, respectively, the AVN classifier utilizes both acoustic and visual information to detect forgeries in videos. 
Table \ref{table8} shows the results of the AV-HuBERT-based AVN classifier on eight test sets. Several observations can be drawn from the table. 
First, the AVN classifier achieves good detection performance on all test sets except the faceswap test set. One possible reason is that the faceswap test set only involves visual manipulation, and the AV-HuBERT feature extractor only extracts visual features from lip images, thereby losing information outside the lip region. 
Second, although the AV-HuBERT-based AVN classifier performs well on the test sets containing manipulation in a single modality . However, its performance may be slightly lower than audio-only and video-only classifiers focusing on one modality. For %example
instance, the AVN classifier achieves an accuracy of 0.96 on RTVC (fake samples containing only acoustic manipulation), but the AN classifier achieves an accuracy of 0.99 on the same test set (see Table \ref{table7}). This is because the AN classifier only focuses on the acoustic modality, while the AVN classifier will inevitably be interfered with by the visual feature extracted from the real visual modality when detecting acoustic manipulation. 
Third, the AV-HuBERT-based AVN classifier achieves accuracy of 0.94 and 0.98 on Testset-I and Testset-II, whose fake instances contain various manipulations in acoustic and visual streams, outperforming the VN classifier (cf. 0.87 and 0.80 in Table \ref{table6}) and the AN classifier (cf. 0.80 and 0.84 in Table \ref{table7}). 
The results reveal that the AVN classifier successfully detects video forgeries by considering both acoustic and visual information. It generally outperforms VN and AN classifiers, which focus on only visual or acoustic manipulation.

\subsubsection{Results of AVTENet}

AVTENet is an ensemble of the above VN, AN, and AVN networks. We compare four fusion strategies, namely majority voting (mv), average score fusion (asf), score fusion (sf), and feature fusion (ff). The resulting models are denoted as $AVTENet_{mv}$, $AVTENet_{asf}$, $AVTENet_{sf}$, and $AVTENet_{ff}$, respectively. AVN is based on AV-HuBERT.

\begin{table}[!t]
% \vspace{-4mm}
\caption{Detection Results of $AVTENet_{asf}$ (Average\\ Score fusion).}
\centering
\scalebox{0.8}{
\begin{tabular}{|c|c|c|c|c|c|c|}
\hline
\textbf {Test-Set type} & \textbf {Class} & \textbf {Precision} & \textbf {Recall} & \textbf {F1-Score} & \textbf {Accuracy} \\
\hline \hline
\multirow{2}{*}{Testset-I} & Real & 0.92 & 1.00  & 0.96 & \multirow{2}{*}{0.96}  \\
\cline{2-5}
& Fake &  1.00 & 0.91 & 0.96 &\\ 
\hline
\multirow{2}{*}{Testset-II} & Real & 1.00 &  1.00 & 1.00 & \multirow{2}{*}{1.00}  \\
\cline{2-5}
& Fake & 1.00 &  1.00 & 1.00 & \\ 
\hline
\multirow{2}{*}{faceswap} & Real & 0.79 & 1.00 & 0.88 & \multirow{2}{*}{0.86} \\
\cline{2-5}
& Fake & 1.00 & 0.73 & 0.84  & \\ 
\hline
\multirow{2}{*}{faceswap-wav2lip} & Real & 1.00 &  1.00 & 1.00 & \multirow{2}{*}{1.00}  \\
\cline{2-5}
& Fake & 1.00 & 1.00 & 1.00 & \\ 
\hline
\multirow{2}{*}{fsgan} & Real & 0.88 & 1.00 & 0.93 & \multirow{2}{*}{0.93}  \\
\cline{2-5}
& Fake & 1.00 & 0.86 & 0.92 & \\ 
\hline
\multirow{2}{*}{fsgan-wav2lip} & Real & 1.00 &  1.00 & 1.00 & \multirow{2}{*}{ 1.00} \\
\cline{2-5}
& Fake &  1.00 &  1.00 & 1.00 & \\ 
\hline
\multirow{2}{*}{RTVC} & Real & 0.96 & 1.00 & 0.98 & \multirow{2}{*}{0.98}  \\
\cline{2-5}
& Fake & 1.00 & 0.96 & 0.98 & \\ 
\hline
\multirow{2}{*}{wav2lip} & Real & 1.00 & 0.99 & 0.99  & \multirow{2}{*}{0.99} \\
\cline{2-5}
& Fake & 0.99 & 1.00 & 0.99 & \\ 
\hline

\end{tabular}}
\label{table10}
\vspace{-6mm}
\end{table}

\begin{table}[!b]
\vspace{-6mm}
\caption{Detection results of $AVTENet_{sf}$ (Score fusion).}
\centering
\scalebox{0.8}{
\begin{tabular}{|c|c|c|c|c|c|c|}
\hline
\textbf {Test-Set type} & \textbf {Class} & \textbf {Precision} & \textbf {Recall} & \textbf {F1-Score} & \textbf {Accuracy} \\
\hline \hline
\multirow{2}{*}{Testset-I} & Real & 0.96 & 0.91 & 0.93 & \multirow{2}{*}{0.94}  \\
\cline{2-5}
& Fake & 0.92 & 0.96 & 0.94 & \\ 
\hline
\multirow{2}{*}{Testset-II} & Real  & 1.00 & 0.91 & 0.96 & \multirow{2}{*}{0.96}  \\
\cline{2-5}
& Fake & 0.92 & 1.00 & 0.96 & \\ 
\hline
\multirow{2}{*}{faceswap} & Real & 0.93 & 0.91 & 0.92 & \multirow{2}{*}{0.92} \\
\cline{2-5}
& Fake & 0.92 & 0.93 & 0.92  & \\ 
\hline
\multirow{2}{*}{faceswap-wav2lip} & Real & 1.00 & 0.91 & 0.96 & \multirow{2}{*}{0.96}  \\
\cline{2-5}
& Fake &  0.92 & 1.00 & 0.96  & \\ 
\hline
\multirow{2}{*}{fsgan} & Real & 0.93 & 0.91 & 0.92 & \multirow{2}{*}{0.92}  \\
\cline{2-5}
& Fake & 0.92 & 0.93 & 0.92 & \\ 
\hline
\multirow{2}{*}{fsgan-wav2lip} & Real & 1.00 &  0.91 &  0.96 & \multirow{2}{*}{ 0.96} \\
\cline{2-5}
& Fake &  0.92 & 1.00 &  0.96 & \\ 
\hline
\multirow{2}{*}{RTVC} & Real & 0.96 & 0.91 & 0.93 & \multirow{2}{*}{0.94}  \\
\cline{2-5}
& Fake & 0.92 & 0.96 & 0.94 & \\ 
\hline
\multirow{2}{*}{wav2lip} & Real & 1.00 & 0.91 & 0.96  & \multirow{2}{*}{0.96} \\
\cline{2-5}
& Fake & 0.92 & 1.00 & 0.96 & \\ 
\hline

\end{tabular}}
\label{table11}
% \vspace{-6mm}
\end{table}

\noindent{$\mathbf{AVTENet_{mv}}$}\textbf{:} 
Table \ref{table9} shows the results of $AVTENet_{mv}$ on eight different test sets. $AVTENet_{mv}$ achieves perfect accuracy of 1.00 on faceswap-wav2lip, fsgan-wav2lip, and Testset-II and high accuracy of 0.99, 0.98, and 0.96 on wav2lip, RTVC, and Testset-I. This result confirms the advantages of integrating VN, AN, and AVN networks in fully considering visual and acoustic information in fake video detection. However, $AVTENet_{mv}$ achieves relatively low performance on faceswap and fsgan. As explained before, this may be because the faceswap and fsgan test sets only involve visual manipulations, while the AV-HuBERT feature extractor ignores visual information outside the lip region. Comparing Table \ref{table9} with Tables \ref{table6}, \ref{table7}, and \ref{table8}, $AVTENet_{mv}$ outperforms all its component networks in all test conditions except for VN on faceswap and AN on RTVC. This result shows that the ensemble $AVTENet_{mv}$ comes with some tradeoffs in fake detection compared to modality-specific classifiers operating on the corresponding unimodal manipulation. %Finally, it is worth mentioning again that $AVTENet_{mv}$ outperforms all its component networks on the two main test sets, Testset-I and Testset-II.

\noindent{$\mathbf{AVTENet_{asf}}$}\textbf{:} 
Table \ref{table10} shows the results of $AVTENet_{asf}$ on eight different test sets. Comparing Table~\ref{table10} and Table~\ref{table9}, we can see that $AVTENet_{asf}$ performs almost the same as $AVTENet_{mv}$, except that the former achieves a slightly lower accuracy than the latter on the faceswap test set (0.86 vs 0.87).

\noindent{$\mathbf{AVTENet_{sf}}$}\textbf{:} 
Table \ref{table11} shows the results of $AVTENet_{sf}$ on eight different test sets. Comparing Table \ref{table11} with Tables \ref{table9} and \ref{table10}, we can see that $AVTENet_{sf}$ performs more stably than $AVTENet_{mv}$ and $AVTENet_{asf}$ across test sets. The stability may be attributed to the additional linear layer taking as input the output scores of the VN, AN, and AVN networks, as it learns to balance the contributions of these three component networks. However, although $AVTENet_{sf}$ outperforms $AVTENet_{mv}$ and $AVTENet_{asf}$ on the faceswap test set (0.92 vs 0.87 and 0.86), it performs worse than $AVTENet_{mv}$ and $AVTENet_{asf}$ on other test sets.

\begin{figure}[!b]
\vspace{-6mm}
\centering
\includegraphics[width=3.1in]{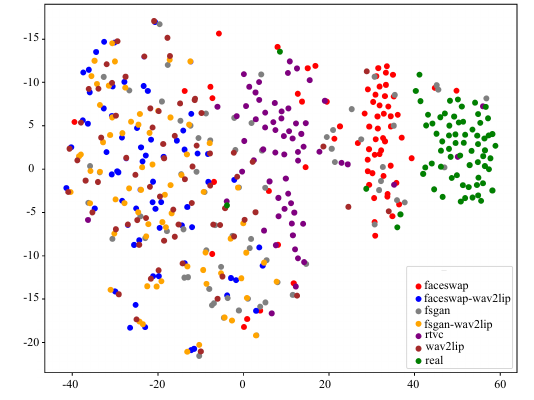}
\vspace{-4mm}
\caption{t-SNE visualization of the fused feature $\mathbf{E_{ff}}$ in $AVTENet_{ff}$.}
\label{Fig6}
% \vspace{-6mm}
\end{figure}

\begin{figure}[!h]
\centering
% \vspace{-3mm}
\includegraphics[width=3.1in]{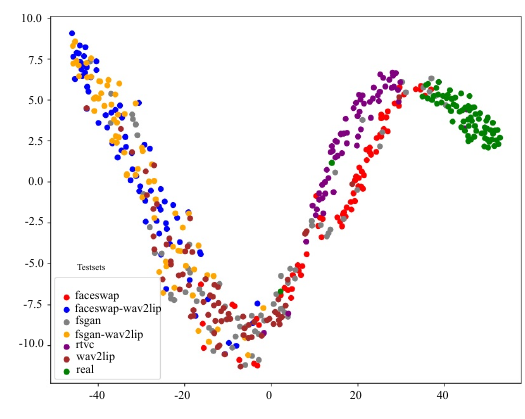}
\vspace{-4mm}
\caption{t-SNE visualization of features in the linear classifier of $AVTENet_{ff}$.}
\label{Fig7}
\vspace{-8mm}
\end{figure}

\noindent{$\mathbf{AVTENet_{ff}}$}\textbf{:} 
Table \ref{table12} shows the results of $AVTENet_{ff}$ on eight different test sets. We can see that $AVTENet_{ff}$ performs best among four ensemble models on faceswap, fsgan, and Testset-II, and achieves near-perfect performance on faceswap-wav2lip, fsgan-wav2lip, wav2lip, and Testset-II. While other ensemble models may perform well on specific test sets, $AVTENet_{ff}$ maintains stable and reliable performance on all eight test sets. The good performance may be attributed to the additional linear layers that takes as input the embeddings of the penultimate layers of the VN, AN, and AVN networks, as it learns to balance the contributions of these three component networks.  
As shown in Figs.~\ref{Fig6} and~\ref{Fig7}, the discriminative ability of the fused feature $\mathbf{E_{ff}}$ is further improved in the additional linear classification layer of $AVTENet_{ff}$. In summary, feature fusion is the best fusion strategy in this study.

\begin{table}[!t]
% \vspace{-6mm}
\caption{Detection Results of $AVTENet_{ff}$ (Feature fusion).}
\centering
\scalebox{0.8}{
\begin{tabular}{|c|c|c|c|c|c|c|}
\hline
\textbf {Test-Set type} & \textbf {Class} & \textbf {Precision} & \textbf {Recall} & \textbf {F1-Score} & \textbf {Accuracy} \\
\hline \hline
\multirow{2}{*}{Testset-I} & Real & 0.96 & 0.97 & 0.96 & \multirow{2}{*}{0.96}  \\
\cline{2-5}
& Fake & 0.97 & 0.96 & 0.96 & \\ 
\hline
\multirow{2}{*}{Testset-II} & Real & 1.00 & 0.97 & 0.99 & \multirow{2}{*}{0.99}  \\
\cline{2-5}
& Fake & 0.97 & 1.00 & 0.99 & \\ 
\hline
\multirow{2}{*}{faceswap} & Real & 0.93 & 0.97 & 0.95 & \multirow{2}{*}{0.95} \\
\cline{2-5}
& Fake & 0.97 & 0.93 & 0.95  & \\ 
\hline
\multirow{2}{*}{faceswap-wav2lip} & Real & 1.00 &  0.97 & 0.99 & \multirow{2}{*}{0.99}  \\
\cline{2-5}
& Fake & 0.97 & 1.00 & 0.99 & \\ 
\hline
\multirow{2}{*}{fsgan} & Real & 0.93 & 0.97 & 0.95 & \multirow{2}{*}{0.95}  \\
\cline{2-5}
& Fake & 0.97 & 0.93 & 0.95 & \\ 
\hline
\multirow{2}{*}{fsgan-wav2lip} & Real & 1.00 &  0.97 &  0.99 & \multirow{2}{*}{ 0.99} \\
\cline{2-5}
& Fake &  0.97 & 1.00 &  0.99 & \\ 
\hline
\multirow{2}{*}{RTVC} & Real & 0.96 & 0.97 & 0.96 & \multirow{2}{*}{0.96}  \\
\cline{2-5}
& Fake & 0.97 & 0.96 & 0.96 & \\ 
\hline
\multirow{2}{*}{wav2lip} & Real & 1.00 & 0.97 & 0.99  & \multirow{2}{*}{0.99} \\
\cline{2-5}
& Fake & 0.97 & 1.00 & 0.99 & \\ 
\hline

\end{tabular}}
\label{table12}
\vspace{-6mm}
\end{table} 

\begin{table*}[!t]
\caption{Performance comparison of  AVTENet (AST\_ViViT\_AST\&ViViT) and AVTENet (AST\_ViViT\_AV-HuBERT).}
    \centering
    \begin{tabular}{|c|c|c|c|c|c|c|c|c|c|}
    \hline
        \textbf{Models} &\textbf{\begin{tabular}{@{}c@{}}Ensemble \\ Strategies\end{tabular} } & \textbf{faceswap} & \textbf{\begin{tabular}{@{}c@{}}faceswap- \\ wav2lip\end{tabular} } & \textbf{fsgan} & \textbf{\begin{tabular}{@{}c@{}}fsgan- \\ wav2lip\end{tabular} } & \textbf{RTVC} & \textbf{wav2lip} & \textbf{Testset-I} & \textbf{Testset-II} \\
        \hline \hline

        \multirow{4}{*}{\textbf{\begin{tabular}[c]{@{}c@{}}AVTENet\\ 
        (AST\_ViViT\_AST\&ViViT)\end{tabular}}} & $AVTENet_{mv}$  & \textbf{0.96} & 0.99 & 0.94 & 0.99  & 0.61 & 0.98 & 0.87 & 0.94 \\
        \cline{2-10}
        & $AVTENet_{asf}$ & \textbf{0.96}   & 0.99  & 0.93 & 0.99  & 0.63  & 0.98   & 0.88   & 0.94   \\
        \cline{2-10}
        & $AVTENet_{sf}$  & 0.91  & 0.93   & 0.88 & 0.93   & 0.52  & 0.91  & 0.80    & 0.87  \\
        \cline{2-10}
        & $AVTENet_{ff}$  & 0.92 & 0.97 & 0.88 & 0.96  & 0.52  & 0.90   & 0.80  & 0.87  \\
        \cline{1-10}
        \hline
        \multirow{4}{*}{\textbf{\begin{tabular}[c]{@{}c@{}}AVTENet\\(AST\_ViViT\_AV-HuBERT)\end{tabular}}}& $AVTENet_{mv}$ & 0.87 & \textbf{1.00}   & 0.93 & \textbf{1.00}    & \textbf{0.98} & \textbf{0.99}  & \textbf{0.96}   & \textbf{1.00}\\
        \cline{2-10}
        & $AVTENet_{asf}$ & 0.86 & \textbf{1.00}   & 0.93 & \textbf{1.00}   & \textbf{0.98}  & \textbf{0.99} & \textbf{0.96}   & \textbf{1.00} \\
        \cline{2-10}
        & $AVTENet_{sf}$  & 0.92 & 0.96    & 0.92 & 0.96  & 0.94 & 0.96 & 0.94  & 0.96 \\
        \cline{2-10}
        & $AVTENet_{ff}$  & 0.95 & 0.99  & \textbf{0.95}  & 0.99  & 0.96  & \textbf{0.99} & \textbf{0.96}  & 0.99  \\       
        \cline{1-10}
        \hline 
    \end{tabular}
    %\vspace{-6mm}
    \label{table13}
\vspace{-6mm}
\end{table*}

\subsubsection{Comparison of The Above Models}
\begin{figure}[!b]
\vspace{-8mm}
\centering
\includegraphics[width=3.3in]{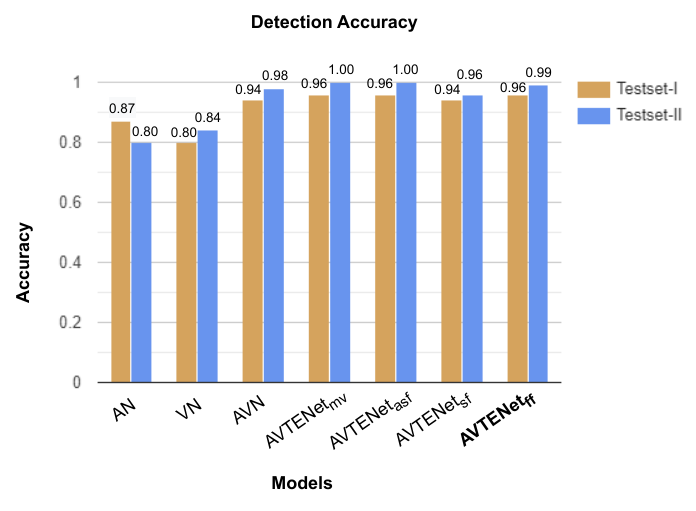}
\vspace{-4mm}
\caption{Performance comparison   of our models on two main test sets, Testset-I and Testset-II.}
\label{Fig8}
% \vspace{-6mm}
\end{figure}

Fig. \ref{Fig8} depicts the accuracy of the above seven models on the two main test sets, Testset-I and Testset-II. The video-only model (VN) is good at identifying visual manipulations in videos. Likewise, the audio-only model (AN) is good at identifying manipulated acoustic content in videos. But they only focus on a single modality. In contrast to VN and AN, the AVN and all variants of AVTENet %are able to 
can identify acoustic and visual manipulations in videos. Although $AVTENet_{mv}$ and $AVTENet_{asf}$ have higher accuracy than $AVTENet_{ff}$ on Testset-II, the latter shows higher stability in detecting various manipulation types in videos. 

Comparing Tables \ref{table9}-\ref{table11} with Table~\ref{table6}, we can see that among the variants of AVTENet, only $AVTENet_{ff}$ can outperform the VN model on all test sets, while the remaining AVTENet models perform worse than the VN model on the faceswap test set (fake samples containing only visual manipulation), despite outperforming the VN model on the other test sets. This is because the VN model only focuses on the visual modality, while the AVTENet models may be interfered with by the acoustic feature extracted from the real auditory modality when detecting visual manipulation. Furthermore, comparing Tables \ref{table9}-\ref{table11} with Table~\ref{table7}, %we find that 
all AVTENet models outperform the AN model on all test sets except RTVC (fake samples containing only acoustic manipulation). %Again, 
This happens because the AN model only focuses on the acoustic modality, while the AVTENet models may be interfered with by the visual feature extracted from the real visual modality when detecting acoustic manipulation.

\begin{table*}[!b]
\vspace{-5mm}
\caption{Performance comparison of our Ensemble model and several existing unimodal, multimodal, fusion, and ensemble models. DFD in the first column refers to the deepfake detection method. ``V'', ``A'' and ``AV'' stand for visual, audio and audio-visual modalities, respectively.}
    \centering
    \begin{tabular}{|c|c|c|c|c|c|c|c|}
    \hline
         \textbf{DFD Method} & \textbf{Model} &  \textbf{Modality} & \textbf{Class} & \textbf{Precision} & \textbf{Recall} & \textbf{F1-score} & \textbf{Accuracy}\\
         \hline\hline
          \multirow{2}{*}{Unimodal\cite{R49}} & \multirow{2}{*}{VGG16} & \multirow{2}{*}{V} & {Real} & 0.69 & 0.90 & 0.78 &  \multirow{2}{*}{0.81} \\
          \cline{4-7} &&& {Fake} & 0.87 & 0.78 & 0.82 & \\
          \hline
          \multirow{2}{*}{Unimodal\cite{R49}} & \multirow{2}{*}{Xception} & \multirow{2}{*}{A} & {Real} & 0.88 & 0.61 & 0.72 & \multirow{2}{*}{0.76} \\
          \cline{4-7} &&& {Fake} & 0.70 & 0.91 & 0.80 & \\
          \hline
          \multirow{2}{*}{Unimodal \cite{R53}} & \multirow{2}{*}{LipForensics} & \multirow{2}{*}{V} & {Real} & 0.70 & 0.91 & 0.80 & \multirow{2}{*}{0.76} \\
          \cline{4-7} &&& {Fake} & 0.88 & 0.61 & 0.72 & \\
          \hline
           \multirow{2}{*}{Ensemble (Soft-Voting)\cite{R49}} & \multirow{2}{*}{VGG16} & \multirow{2}{*}{AV} & {Real} & 0.69 & 0.90 & 0.78 & \multirow{2}{*}{0.78} \\
          \cline{4-7} &&& {Fake} & 0.90 & 0.69 & 0.78 & \\
          \hline
           \multirow{2}{*}{Ensemble (Hard-Voting)\cite{R49}} & \multirow{2}{*}{VGG16} & \multirow{2}{*}{AV} & {Real} & 0.69 & 0.90 & 0.78 & \multirow{2}{*}{0.78} \\
          \cline{4-7} &&& {Fake} & 0.90 & 0.69 & 0.78 & \\
          \hline
          \multirow{2}{*}{Multimodal-1\cite{R49}} & \multirow{2}{*}{Multimodal-1} & \multirow{2}{*}{AV} & {Real} & 0.00 & 0.00 & 0.00& \multirow{2}{*}{0.50} \\
          \cline{4-7} &&& {Fake} & 0.50 & 1.00 & 0.66 & \\ 
          \hline
          \multirow{2}{*}{Multimodal-2\cite{R49}} & \multirow{2}{*}{Multimodal-2} & \multirow{2}{*}{AV} & {Real} & 0.71 & 0.59 & 0.64 & \multirow{2}{*}{0.67} \\
          \cline{4-7} &&& {Fake} & 0.65 & 0.76 & 0.70 & \\
          \hline
          \multirow{2}{*}{Multimodal-3\cite{R49}} & \multirow{2}{*}{CDCN} & \multirow{2}{*}{AV} & {Real} & 0.50 & 0.07 & 0.12 & \multirow{2}{*}{0.52} \\
          \cline{4-7} &&& {Fake} & 0.50 & 0.94 & 0.65 & \\
          \hline
          \multirow{2}{*}{Multimodal-4\cite{R34}} & \multirow{2}{*}{Not-made-for-each-other} & \multirow{2}{*}{AV} & {Real} & 0.62 & 0.99 & 0.76 & \multirow{2}{*}{0.69} \\
          \cline{4-7} &&& {Fake} & 0.94 & 0.40 & 0.57 & \\
          \hline
          \multirow{2}{*}{{Multimodal \cite{R6}}} & \multirow{2}{*}{{ Ensemble}} & \multirow{2}{*}{{AV}} &{Real} & {0.83} & {0.99} &  {0.90} & \multirow{2}{*}{{0.89}} \\
          \cline{4-7} &&& { {Fake}} &  {0.98} &  {0.80} & {0.88} & \\
          \hline
          \multirow{2}{*}{{Multimodal \cite{R20}}} & \multirow{2}{*}{{ AV-Lip-Sync}} & \multirow{2}{*}{{AV}} &{Real} & {0.93} & {0.96} &  {0.94} & \multirow{2}{*}{{0.94}} \\
          \cline{4-7} &&& { {Fake}} &  {0.96} &  {0.93} & {0.94} & \\
          \hline
          \multirow{2}{*}{Modality Mixing \cite{R50}} & \multirow{2}{*}{ Multimodaltrace } & \multirow{2}{*}{AV} & {Real} &  {-} & {-} & {-} & \multirow{2}{*}{0.93} \\
          \cline{4-7} &&& {Fake} &  {-} &  {-} & {-} & \\
          \hline
          \multirow{2}{*}{ Ensemble\cite{R51}} & \multirow{2}{*}{ AVFakeNet } & \multirow{2}{*}{AV} & {Real} &  {-} & {-} & {-} & \multirow{2}{*}{0.93} \\
          \cline{4-7} &&& {Fake} &  {-} &  {-} & {-} & \\
          \hline
          \multirow{2}{*}{ Fusion\cite{R52}} & \multirow{2}{*}{ AVoiD-DF } & \multirow{2}{*}{AV} & {Real} & {-} & {-} & {-} & \multirow{2}{*}{0.84} \\
          \cline{4-7} &&& {Fake} &  {-} &  {-} & {-} & \\
          \hline
          \multirow{2}{*}{ Fusion\cite{R59}} & \multirow{2}{*}{ MIS-AVioDD } & \multirow{2}{*}{AV} & {Real} & {-} & {-} & {-} & \multirow{2}{*}{0.96} \\
          \cline{4-7} &&& {Fake} &  {-} &  {-} & {-} & \\
          \hline
          \multirow{2}{*}{\textbf{Ensemble (ours)}} & \multirow{2}{*}{$\mathbf{AVTENet_{ff}}$} & \multirow{2}{*}{AV} & Real & \textbf {1.00} & \textbf {0.97} & \textbf {0.99} & \multirow{2}{*}{\textbf {0.99}} \\
          \cline{4-7} &&& {Fake} & \textbf {0.97} & \textbf {1.00} & \textbf {0.99} & \\
          \hline
    \end{tabular}
    \label{table14}
\vspace{-4mm}
\end{table*}

\subsubsection{Comparison of AVTENet with Different AVNs}
Our proposed AVTENet model is an ensemble of an AST-based AN model, a ViViT-based VN model, and an AV-HuBERT-based AVN model. Although $AVTENet_{ff}$ performs well on all test sets, its performances on the faceswap and fsgan test sets are relatively low, where the fake instances are only visually manipulated while
audio tracks are real. We speculate that the reason may be that the AV-HuBERT feature extractor in AVN only extracts visual information from lip images and ignores information outside the lip region. Therefore, we also implement another AVN based on AST and ViViT. Here, we compare AVTENet (AST\_ViViT\_AV-HuBERT) and AVTENet (AST\_ViViT\_AST\&ViViT). From Table \ref{table13}, %it is clear that 
although the AST\_ViViT\_AST\&ViViT-based $AVTENet_{mv}$ and $AVTENet_{asf}$ models do slightly improve the performance on the faceswap test set, the AST\_ViViT\_AST\&ViViT-based AVTENet models perform worse than the AST\_ViViT\_AV-HuBERT-based AVTENet models. %The reason may be that since 
The AST and ViViT models are already used in the AN and VN models. Therefore, they cannot provide additional information on the AVN model. Surprisingly, the AST\_ViViT\_AST\&ViViT-based AVTENet models almost fail on the RTVC test set. To fully utilize face images in the AVN model, we may need to apply new pre-trained visual models. 

\begin{figure}[!t]
% \vspace{-8mm}
\centering 
\includegraphics[width=3.5in]{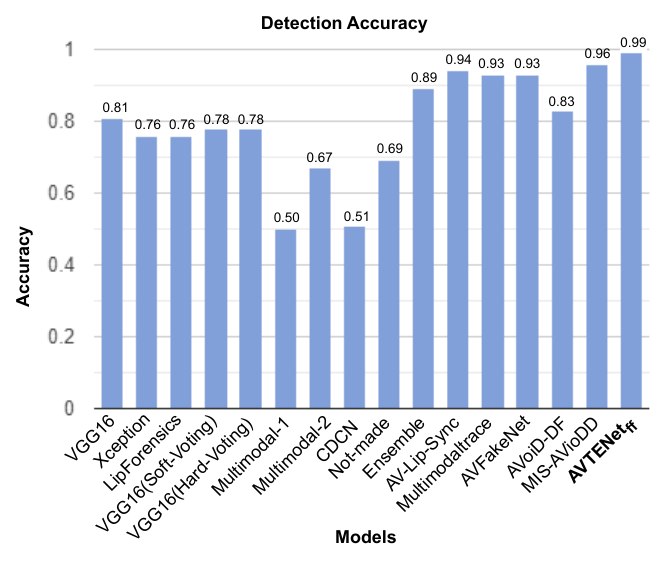}
\vspace{-8mm}
\caption{The comparison of our proposed approach ($AVTENet_{ff}$) with other existing unimodal, multimodal, fusion, and ensemble benchmarks.}
\label{Fig9}
\vspace{-6mm}
\end{figure} 

\subsubsection{Comparison of AVTENet and Other Models}

Finally, as shown in Fig. \ref{Fig9} and Table \ref{table14}, we conduct a thorough comparison of our proposed $AVTENet_{ff}$ model with various existing unimodal, multimodal, fusion, and ensemble methods. Several unimodal, multimodal and ensemble methods were presented in \cite{R49}, but most of them failed to detect acoustic and visual manipulations in videos well. VGG16, a unimodal approach exclusively trained on visual data, outperforms all other unimodal approaches reported in \cite{R49} for video forgery detection with an accuracy of 0.81. Xception, another unimodal method reported in \cite{R49}, is trained exclusively on the audio modality and achieves an accuracy of only 0.76. Similarly, Lip Forensics \cite{R17} exploits visual lip movements to detect forgeries in videos with an accuracy of 0.76. Although Ensemble (soft-voting and hard-voting), Multimodal-1, Multimodal-2, and Multimodal-3 utilize both audio and visual modalities for deepfake detection, their performance is still unsatisfactory. Not-made-for-each other \cite{R34} is an audio-visual dissonance-based approach that detects video forgeries with an accuracy of 0.69. The multimodal ensemble approach proposed in \cite{R6} utilizes three CNN-based classifiers to identify forgeries and achieves an accuracy of 0.89. Another audio-visual approach proposed in \cite{R20} successfully identifies fake videos using lip synchronization, and the model achieves an accuracy of 0.94. Moreover, the ensemble approach Multimodaltrace\cite{R50} %is an ensemble approach proposed in \cite{R50}, which 
achieves an accuracy of 0.93. AVFakeNet\cite{R51} %\cite{R51} 
is also an ensemble approach using audio and video transformer encoders, which achieves an accuracy of 0.93. AVoiD-DF \cite{R52} is another audio-visual fusion approach, but it only achieves an accuracy of 0.84.  Furthermore, MIS-AVioDD\cite{R59} %\cite{R59} 
is also a fusion approach that jointly uses the modality invariant and modality-specific representations to detect audio-visual forgery and achieves an accuracy of 0.96. %From
Table \ref{table14} shows %it is evident 
that our ensemble approach outperforms other existing models, %achieving 
with an accuracy of 0.99 on Testset-II of the FakeAVCeleb dataset, which is a new SOTA performance.

\begin{table}[!b]
\vspace{-8mm}
\caption{Performance comparison of the proposed model and other models tested on the DFDC test set.}
% \vspace{-4mm}
\begin{center}
\scalebox{0.9}{
\resizebox{\columnwidth}{!}{%
\begin{tabular}{|c|c|c|}
\hline \textbf{Model} & \textbf{AUC Score (\%)} \\
\hline
\hline LipForensics \cite{R53} & 53.10 \\
\hline CViT \cite{R70} & 56.70 \\
\hline MesoNet \cite{R24} & 41.20 \\
\hline MDS \cite{R34} & 73.80 \\
\hline AVoiD-DF \cite{R52} & 80.60  \\
\hline Proposed model ($\text{AVTENet}_\text{ff}$) & 64.10 \\
\hline Proposed model ($\text{AVTENet}_\text{ff}$) (on the frontal face subset) & 83.84 \\
\hline
\end{tabular}}
}
\label{DFDC_res}
\end{center}
\vspace{-2mm}
\end{table}

\subsubsection{Comparison of AVTENet and Humans} 
We conducted a preliminary comparison of AVTENet’s detection capabilities with human cognitive proficiency in detecting audio-visual deepfakes using a manually curated subset of 30 videos (15 real videos and 15 fake videos) from the FakeAVCeleb test set to ensure a diverse and challenging evaluation set, with all fake videos featuring audio-visual manipulation.

The videos were presented to 24 subjects in three phases to classify each video as real or fake. In the first phase (audio-only), participants judged whether the video was authentic or not by listening to the audio track alone. In the second phase (video-only), participants judged the authenticity of the video using only visual input. Finally, in the third phase (audio-visual), participants integrated the audio and visual streams to classify the video as real or fake. As shown in Table~\ref{table:evaluation}, humans rely on their own perception and reasoning capabilities to achieve an accuracy of 0.58 under the audio-only condition, 0.63 under the video-only condition, and 0.76 under the audio-visual condition.
From Table~\ref{table:evaluation}, it is clear that, under three conditions, AST (audio-only network), ViViT (video-only network), and AVTENet (audio-visual network) outperform humans (0.77 vs 0.58, 0.83 vs 0.63, and 0.93 vs 0.76), respectively.
The results highlight the importance of multimodal sensory integration for enhancing human and model performance in video forgery detection; especially for detection models, combining audio and visual cues can lead to significant performance improvements.

\begin{table}[h!]
\vspace{-3mm}
\centering
\caption{Comparison of humans and models in video forgery detection.}
\begin{tabular}{|c|c||c|c|}
\hline 
\multicolumn{2}{|c||}{\textbf{Humans}} & \multicolumn{2}{c|}{\textbf{Models}} \\ \hline \hline
\textbf{Modalities} & \textbf{{\begin{tabular}[c]{@{}c@{}} Accuracy\end{tabular}}} & \textbf{Modalities} & \textbf{\begin{tabular}[c]{@{}c@{}}Accuracy\end{tabular}} \\ 
        \hline
Audio-Only & 0.58 & AST (Audio-Only) & 0.77 \\ \hline
Video-Only & 0.63 & ViViT (Video-Only) & 0.83 \\ \hline
Audio-visual & 0.76 & AVTENet (Audio-visual) & 0.93 \\ \hline
\end{tabular}
\label{table:evaluation}
\vspace{-2mm}
\end{table}

\subsubsection{Generalization Test}
Generalization remains a significant challenge for video forgery detection tasks, including multimodal deepfake detection. To evaluate the generalization ability of our proposed method, we directly test the proposed model trained on the FakeAVCeleb dataset on the DFDC test set, and compare it with other models under the same setting. 
We extract the test videos and preprocess them following \cite{R34} and \cite{R71}.
As shown in Table~\ref{DFDC_res}, our model achieves an AUC (area under the receiver operating characteristic curve) score of 64.10\%, which is better than the performance of LipForensics \cite{R53}, CViT \cite{R70}, and MesoNet \cite{R24}, but worse than the performance of MDS \cite{R34} and AVoiD-DF \cite{R52}.
The superior performance of these methods may be due to the fact that they focus on the entire image frame rather than just the frontal face or lip regions. 

Our study focuses on developing a multimodal deepfake detection method for frontal face videos. However, the DFDC test set contains various types of deepfake videos, including distant faces, side faces, multiple speakers, illumination variations, and low-light conditions. This variation poses challenges to the accurate extraction of faces and lips. Therefore, we took the frontal face videos from the DFDC test set and created a frontal face video test subset. As shown in Table~\ref{DFDC_res}, our model achieves an AUC score of 83.84\% on the frontal face video test subset. The results show that our proposed model remains highly effective for frontal face videos where facial identity or expression as well as lip movements may be manipulated, and shows reasonable performance on the entire test set. The results also show that our model has quite good generalization ability. 

\section{Limitations}
This %work 
study focuses on the authenticity detection of frontal face videos. Lack of frontal face, extreme environmental conditions, illumination variation, and multiple speakers in videos pose challenges to the face detection and lip capture modules in our method. For such types of videos, complete frame-based image analysis is required. However, in the proposed ensemble approach, the video-only model analyzes visual inconsistencies in facial movements, lighting, and other video content, independent of lip image data. While utilizing lip image data, the audio-visual model supplements the other two models by cross-referencing audio-visual synchronization to improve detection accuracy. Furthermore, our method is bimodal, requiring the input to contain both visual and acoustic modalities. To train our model, bimodal manipulations are required to train independent modality-specific networks and integrate their features into the decision-making module. 

Due to the lack of suitable audio-visual datasets, it is currently not possible to test the generalization ability of the proposed model on more datasets. In addition to awaiting new publicly available audio-visual deepfake datasets, we are also preparing to expand the DeepfakeTIMIT dataset~\cite{R63}, where the audio in both real and fake videos is %always 
consistently authentic. We %plan
aim to apply various cutting-edge voice conversion technologies for acoustic manipulation. 

As our ensemble approach consists of multiple transformers, the overall computational cost is higher compared to CNN-based or single transformer-based models in the literature. The number of trainable parameters in our video-only, audio-visual, and audio-only networks are 114.75M, 129.31M, and 87.73M respectively. The number of parameters in the linear layers of the decision-making module is 2.89M. In total, our proposed AVTENET model contains 334.68M trainable parameters. For inference time, we evaluate our proposed model on an NVIDIA GeForce RTX 2080 Ti GPU. To ensure accuracy, we executed the proposed model in the evaluation mode and performed multiple warm-up iterations before measurement to eliminate cold start overhead. The average inference time per video sample is 0.55 seconds. The recorded inference time only includes the forward pass of the proposed model and does not include  any pre- or post-processing time.

\section{Conclusions and Future Work}
In this study, we demonstrated the effectiveness of ensemble learning and studied deepfake video detection from an audio-visual perspective. We proposed a novel audio-visual transformer-based ensemble learning solution for effective and scalable video forgery detection. This approach is particularly effective for 
videos containing acoustic and visual manipulations, which cannot rely on existing methods because they can only detect unimodal manipulations or lack training diversity. Specifically, we devised a transformer-based model with a powerful detection ability due to its sequence modeling, parallelization, and attention mechanisms. We illustrated the performances of our multimodal transformer-based ensemble model on several manipulation techniques and compared 
them with various existing unimodal, multimodal, fusion, and ensemble models. The proposed model achieved state-of-the-art performance on the FakeAVCeleb dataset. To enhance trust and security in digital media environments, our model can be used to instantly detect deepfakes on social media platforms, which aids in content moderation and forensic analysis. Furthermore, the proposed solution can be extended through various foundation models trained using large amounts of data.

\section*{Acknowledgments}
This work was supported in part by Academia Sinica under Grant AS-GC-111-M01, and by the National Council of Science and Technology, Taiwan, under Grants NSTC 113-2634-F-002-003 and NSTC 112-2221-E-001-009-MY3. 

\bibliographystyle{IEEEtran}
\bibliography{Reference}

% \vspace{11pt}

\begin{IEEEbiography}
[{{\includegraphics[width=1in,height=1.25in,clip,keepaspectratio]{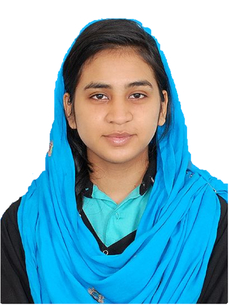}}}]{Ammarah Hashmi} received her B.S. in Software Engineering from COMSATS University, Pakistan in 2017, followed by an M.S. in Software Engineering from Beijing Institute of Technology, China in 2019, where she specialized in gamification and its applications in user engagement and digital experiences. She is currently pursuing her Ph.D. in Social Networks and Human-Centered Computing at the Institute of Information Systems and Applications, National Tsing Hua University, Taiwan, through a collaborative program with the Institute of Information Science, Academia Sinica, Taiwan. Her research focuses on deep learning-based forgery detection, multimodal data processing, multimedia forensics, and AI-driven content authentication. She has contributed to several projects related to image and video manipulation detection, deepfake analysis, and digital forensic methodologies.
\end{IEEEbiography}
\vspace{-10pt} 
\begin{IEEEbiography}
[{{\includegraphics[width=1in,height=1.25in,clip,keepaspectratio]{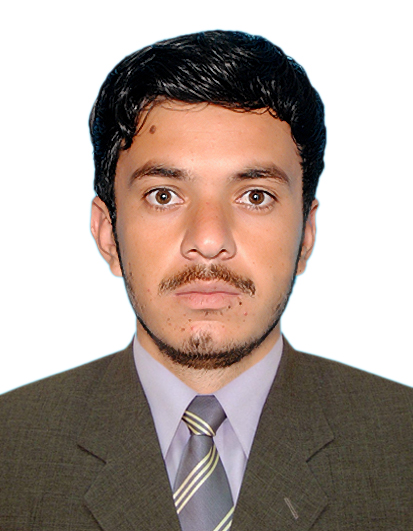}}}]{Sahibzada Adil Shahzad} received his B.S. degree in Software Engineering from the University of Science and Technology Bannu, Pakistan, in 2016, and his M.S. degree in Computer Science from the Beijing Institute of Technology, China, in 2019. He is currently pursuing a Ph.D. in the Taiwan International Graduate Program in Social Networks and Human-Centered Computing, jointly offered by the Institute of Information Science, Academia Sinica, and the Department of Computer Science, National Chengchi University, Taiwan. During his bachelor's and master's studies, he worked on mobile app development and mobile augmented reality applications leveraging deep learning techniques. His current research focuses on machine learning, multimedia forensics, and audio-visual processing, with an emphasis on deepfake detection and digital media authentication.
\end{IEEEbiography}

\begin{IEEEbiography}
[{\includegraphics[width=1in,height=1.25in,clip,keepaspectratio] {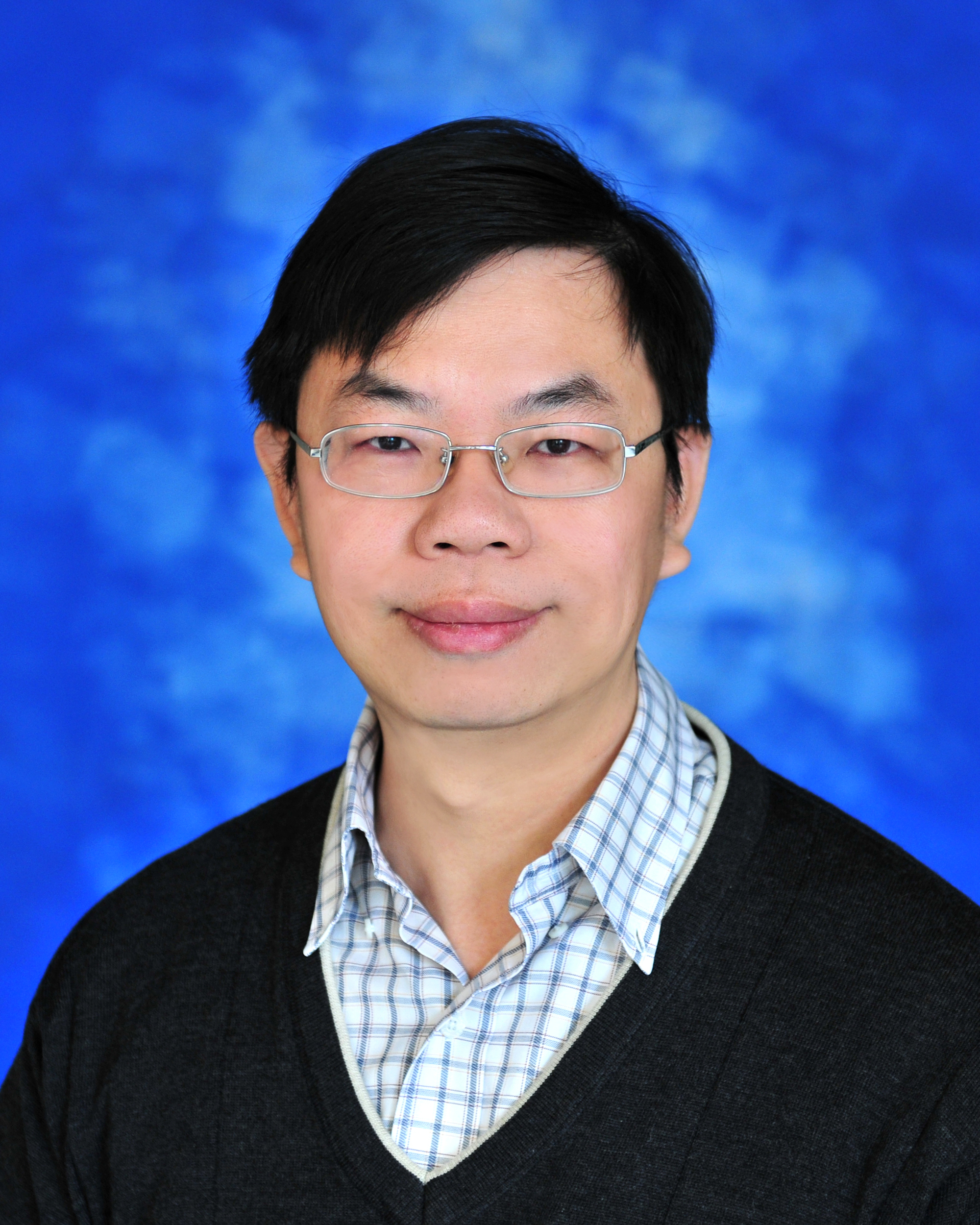}}]
{Chia-Wen Lin}
(Fellow, IEEE) received his Ph.D. degree from National Tsing Hua University (NTHU), Hsinchu, Taiwan, in 2000.  
Dr. Lin is currently a Distinguished Professor with the Department of Electrical Engineering and the Institute of Communications Engineering, NTHU. His research interests include image/video processing and computer vision.  He has served as a Fellow Evaluation Committee member (2021--2023), BoG Members-at-Large (2022--2024), and Distinguished Lecturer (2018--2019) of the IEEE Circuits and Systems Society.   He was Chair of IEEE ICME Steering Committee (2020--2021). He served as TPC Co-Chair of PCS 2022, IEEE ICIP 2019 and IEEE ICME 2010, and General Co-Chair of PCS 2024 and IEEE VCIP 2018.  He received two best paper awards from VCIP 2010 and 2015. He was an Associate Editor of \textsc{IEEE Transactions on Image Processing}, \textsc{IEEE Transactions on Circuits and Systems for Video Technology}, \textsc{IEEE Transactions on Multimedia}, and \textsc{IEEE Multimedia}. He received the Distinguished Research Award presented by National Science and Technology Council, Taiwan, in 2023, and IEEE VCIP Best Paper Award in 2015.
\end{IEEEbiography}

\begin{IEEEbiography}[{{\includegraphics[width=1in,height=1.25in,clip,keepaspectratio]{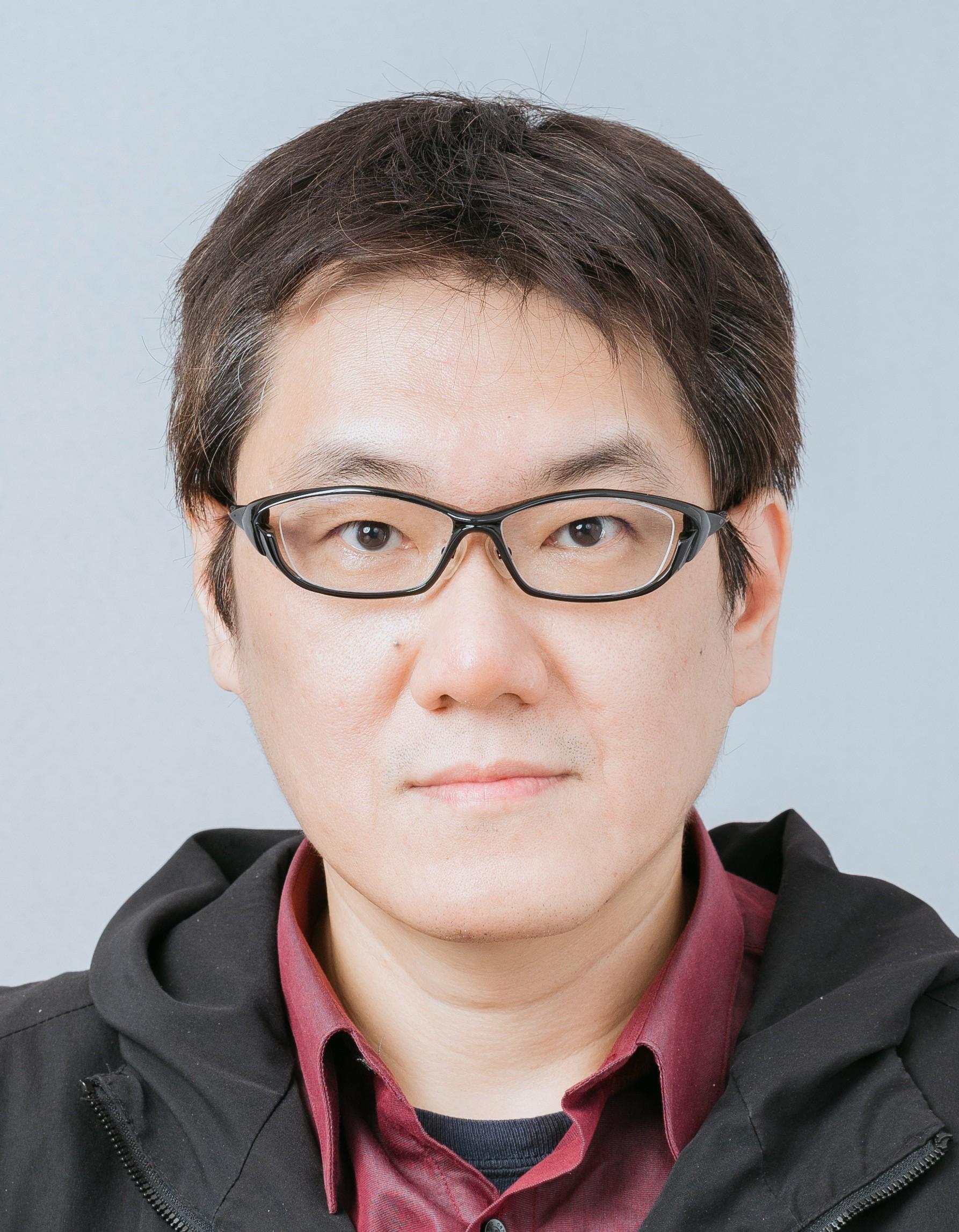}}}]{Yu Tsao} Yu Tsao (Senior Member, IEEE) received his B.S. and M.S. degrees in Electrical Engineering from National Taiwan University, Taipei, Taiwan, in 1999 and 2001, respectively, and his Ph.D. degree in Electrical and Computer Engineering from the Georgia Institute of Technology, Atlanta, GA, USA, in 2008. From 2009 to 2011, he was a Researcher at the National Institute of Information and Communications Technology, Tokyo, Japan, where he worked on research and product development for automatic speech recognition in multilingual speech-to-speech translation. He is currently a Research Fellow (Professor) and the Deputy Director at the Research Center for Information Technology Innovation, Academia Sinica, Taipei, Taiwan. Additionally, he serves as a Jointly Appointed Professor in the Department of Electrical Engineering, Chung Yuan Christian University, Taoyuan, Taiwan. Dr. Tsao’s research interests include assistive oral communication technologies, audio coding, and bio-signal processing. He is an Associate Editor for both IEEE Transactions on Consumer Electronics and IEEE Signal Processing Letters. In recognition of his contributions, he received the Outstanding Research Award from the National Science and Technology Council (NSTC)—the most prestigious research honor in Taiwan—in 2023. His other accolades include the Academia Sinica Career Development Award in 2017, multiple National Innovation Awards (2018–2021 and 2023), the Future Tech Breakthrough Award in 2019, the Outstanding Elite Award from the Chung Hwa Rotary Educational Foundation (2019–2020), and the NSTC FutureTech Award in 2022. Additionally, he is the corresponding author of a paper that won the 2021 IEEE Signal Processing Society (SPS) Young Author Best Paper Award.
\end{IEEEbiography}

\begin{IEEEbiography}[{{\includegraphics[width=1in,height=1.25in,clip,keepaspectratio]{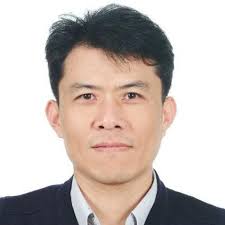}}}]{Hsin Min Wang}(Senior Member, IEEE) received the B.S. and Ph.D. degrees in electrical engineering from National Taiwan University, Taipei, Taiwan, in 1989 and 1995, respectively. In October 1995, he joined the Institute of Information Science, Academia Sinica, Taipei, Taiwan, where he is currently a Research Fellow. He was a Joint Professor in the Department of Computer Science and Information Engineering at National Cheng Kung University from 2014 to 2023. He was an Associate Editor of IEEE/ACM Transactions on Audio, Speech and Language Processing from 2016 to 2020. He currently serves a Senior Editor of APSIPA Transactions on Signal and Information Processing. His major research interests include spoken language processing, natural language processing, multimedia information retrieval, and machine learning. He was a General Co-Chair of ISCSLP2016, ISCSLP2018, and ASRU2023 and a Technical Co-Chair of ISCSLP2010, O-COCOSDA2011, APSIPAASC2013, ISMIR2014, ASRU2019, and APSIPAASC2023. He received the Chinese Institute of Engineers Technical Paper Award in 1995 and the ACM Multimedia Grand Challenge First Prize in 2012. He was an APSIPA distinguished lecturer for 2014–2015. He is a member of ISCA, ACM, and APSIPA.
\end{IEEEbiography}
% \vspace{11pt}

\vfill

\end{document}